\documentclass{article} 
\usepackage{iclr2027_conference,times}

\usepackage[utf8]{inputenc}
\usepackage[T1]{fontenc}

\usepackage{amsmath,amsfonts,bm}

\def\Figref#1{Figure~\ref{#1}}

\def\Secref#1{Section~\ref{#1}}

\def\eqref#1{equation~\ref{#1}}

\def\1{\bm{1}}

\DeclareMathAlphabet{\mathsfit}{\encodingdefault}{\sfdefault}{m}{sl}
\SetMathAlphabet{\mathsfit}{bold}{\encodingdefault}{\sfdefault}{bx}{n}

\usepackage{graphicx}
\usepackage{subcaption}

\usepackage{amsmath}
\usepackage{amssymb}
\usepackage{amsfonts}

\usepackage{booktabs}
\usepackage{multirow}
\usepackage{tabularx}

\usepackage{microtype}

\usepackage{wrapfig}
\usepackage{enumitem}
\setlist[itemize]{nosep, leftmargin=1.5em}
\usepackage{xcolor}

\usepackage{hyperref}
\usepackage{url}

\newcommand{\Tabref}[1]{Table~\ref{#1}}
\newcommand{\Appref}[1]{Appendix~\ref{#1}}

\title{After a Decade: Bringing Shadow Removal into the Real World with Agentic Training Data}

\author{
Shilin Hu\textsuperscript{1} \quad
Jingyi Xu\textsuperscript{1} \quad
Dimitris Samaras\textsuperscript{1}\thanks{Equal advising.} \quad
Hieu Le\textsuperscript{2}\footnotemark[1]
\\[0.5em]
\textsuperscript{1}\begin{tabular}[t]{@{}l@{}}
Department of Computer Science\\
Stony Brook University
\end{tabular}
\qquad
\textsuperscript{2}\begin{tabular}[t]{@{}l@{}}
Department of Computer Science\\
University of North Carolina at Charlotte
\end{tabular}
\\[0.5em]
{\small
\texttt{\{shilhu,jingyixu,samaras\}@cs.stonybrook.edu}
\quad
\texttt{hle40@charlotte.edu}
}
}

\iclrfinalcopy 
\begin{document}

\maketitle

\fancyhead{}
\renewcommand{\headrulewidth}{0pt}

\begin{abstract}

Shadow removal looks nearly solved on established benchmarks, yet remains brittle in the real world. Models have advanced; the paired training data they rely on have barely changed in nearly a decade. The reason is simple: obtaining a shadow-free target requires removing the occluder while keeping the scene, camera, and illumination otherwise unchanged, making diverse paired data difficult to capture. Meanwhile, large shadow detection datasets already contain diverse real-world images and masks, but no shadow-free targets. To turn this abundant but incomplete data into paired supervision, we propose an offline agentic workflow combining physics-motivated generation, failure detection, feedback-driven retry, candidate selection, and deterministic correction. Using this workflow, we construct \textbf{AgenticShadow}, a dataset of 17,138 image-mask-target triplets spanning general scenes, faces, and remote sensing. Our construction workflow reduces Color Distribution Difference by 50.5\% over previous shadow removal work, while training existing shadow removal models on AgenticShadow reduces cross-domain LAB RMSE by 19.7--37.5\%.

\end{abstract}

\section{Introduction}

Shadow removal appears increasingly mature on standard paired benchmarks. On ISTD+~\citep{wang2018stacked,le2019shadowdecomposition}, several recent methods report relatively small differences among leading approaches with visually near-perfect results~\citep{guo2023shadowdiffusion,xiao2024homoformer,xu2025detail,lee2026phasr}. Yet this performance is limited to the relatively simple cases represented in the dataset and does not carry over reliably to diverse real-world images. Large cast shadows in outdoor scenes, mixed illumination in street imagery, shadows across faces, or aerial scenes with broad occlusions frequently lead to incomplete removal, color shifts, and structural artifacts. The reason is simple: the conditions encountered in practice are much broader than the paired data available for training.

Collecting a valid pair requires removing the shadow while keeping the camera, scene geometry, materials, exposure, and other illumination conditions unchanged. This is manageable in controlled capture, but difficult to scale to streets, faces, complex outdoor scenes, or aerial imagery. SRD~\citep{qu2017deshadownet} and ISTD~\citep{wang2018stacked} remain the principal paired datasets used for shadow removal. Later efforts improve different aspects of this data: ISTD+~\citep{le2019shadowdecomposition} corrects color inconsistencies, while SRD+~\citep{inoue2021learning} cleans and re-splits SRD. WSRD and WSRD+~\citep{vasluianu2023wsrd,vasluianu2024ntire} provide high-resolution paired imagery through calibrated, close-range indoor capture. Nevertheless, paired acquisition still constrains both scale and domain coverage. The bottleneck is therefore not simply obtaining more shadow images, but obtaining their corresponding shadow-free targets under sufficiently varied conditions.

The main contribution of this work is to bridge this decade-old data gap. We do so with a physics-motivated agentic workflow using multimodal foundation models that converts real shadow image-mask pairs into paired training data. Rather than accepting a single prompted edit, the workflow treats generation as a proposal: it evaluates an initial candidate, detects major failures, triggers feedback-guided retries when needed, filters invalid outputs, and selects among valid candidates using visual and structural cues. A final deterministic regional correction reduces residual photometric inconsistencies. This generate--evaluate--retry--select--correct loop produces pseudo shadow-free targets at scale while explicitly controlling the failure modes of generative editing. This process yields \textbf{AgenticShadow}: 17,138 image--mask--target triplets spanning general scenes, faces, and remote sensing. Crucially, the resulting dataset is not simply larger. It expands the distribution of paired supervision to larger and more spatially distributed shadows, lower-contrast cases, and substantially more heterogeneous correction requirements than established paired benchmarks.

\begin{figure}[!t]
\centering
\includegraphics[width=\linewidth]{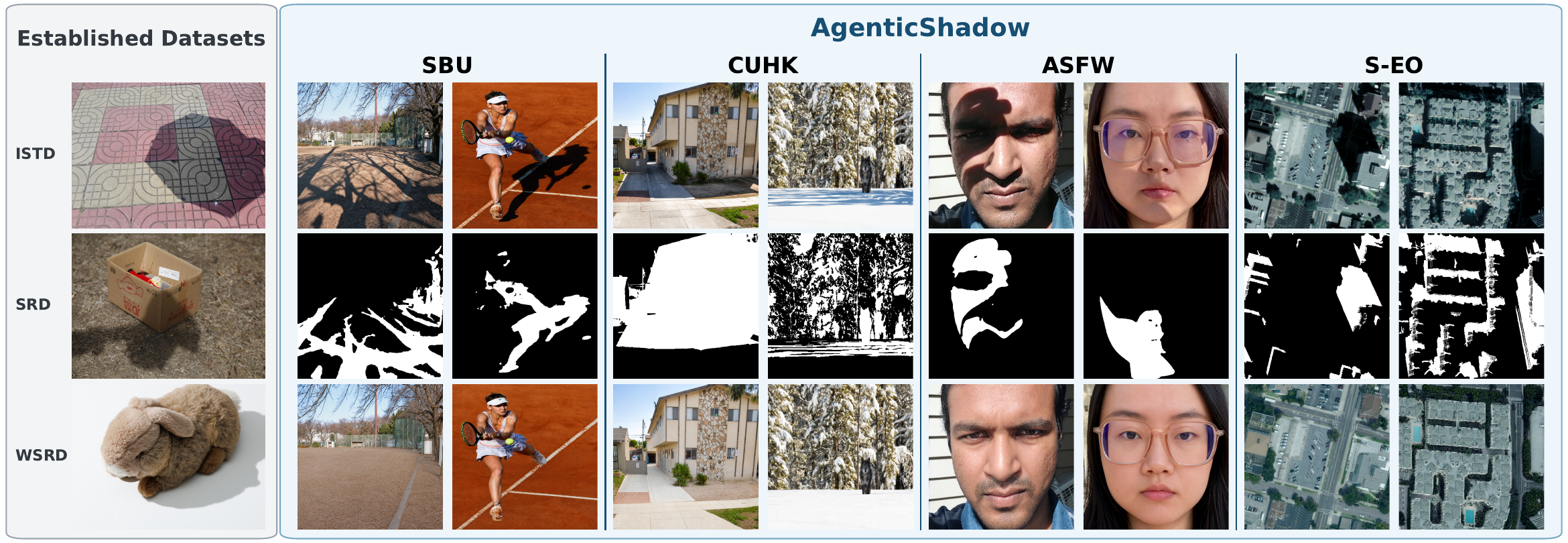}
\caption{\textbf{From constrained paired capture to diverse real-world supervision.}
ISTD~\citep{wang2018stacked} and SRD~\citep{qu2017deshadownet} largely reflect constrained paired acquisition with limited scene diversity, while WSRD~\citep{vasluianu2023wsrd} remains centered on controlled, close-range object and surface interactions. AgenticShadow draws from SBU~\citep{vicente2016largescale}, CUHK-Shadow~\citep{hu2021revisiting}, ASFW~\citep{luo2026beyond}, and S-EO~\citep{masquil2025seo}, providing substantially broader scenes, viewpoints, shadow appearances, and application domains.}
\label{fig:teaser}
\vspace{-2mm}
\end{figure}

Simply put, our proposed AgenticShadow dataset provides supervision for shadow distributions where paired capture is impractical. Our pseudo targets are not physical ground truth. In unconstrained scenes, however, an exact shadow-free counterpart may be inherently unobtainable: one cannot remove a building to eliminate its shadow for data collection, for example. Even conventional paired capture~\citep{wang2018stacked} provides only an approximation, since removing or repositioning an occluder can perturb illumination, alignment, or exposure. Our goal is therefore not exact physical reconstruction, but a controlled approximation that preserves scene content while restoring illumination. We enforce this through shadow-formation grounding in generation and evaluation, rejection of candidates that alter scene content or retain shadow-like structure, structural checks for added or lost detail, and regional photometric correction for residual brightness and color errors. These constraints do not guarantee a unique physical solution, but they make the constructed targets sufficiently controlled to serve as supervision. As we show in \Secref{sec:agentic_quality}, the ablations support this design directly: adding these controls progressively improves target quality. Under generic prompting, retry and selection already reduce Color Distribution Difference (CDD), which compares RGB color distributions sampled along the shadow and non-shadow sides of annotated boundaries. Shadow-formation grounding improves every corresponding stage, and structural diagnostics and regional correction further reduce the final error.

We validate both the fidelity of the constructed targets and their value as training data. Without training on ISTD+ or access to the references during construction, our agentic workflow produces targets that closely match the color-corrected ISTD+ ground truth, achieving a LAB RMSE of 3.85. On the benchmark introduced in ShadowRemovalRefine~\citep{hu2025shadowrefine}, the workflow reduces CDD by 50.5\% relative to the strongest previous method. More importantly, this supervision transfers across architectures, reducing cross-domain macro LAB RMSE by 19.7--37.5\% for ShadowDiffusion~\citep{guo2023shadowdiffusion}, HomoFormer~\citep{xiao2024homoformer}, and PhaSR~\citep{lee2026phasr}.

Our main contributions are:
\begin{itemize}

\item We introduce AgenticShadow, to our knowledge the largest paired shadow removal benchmark constructed from diverse real-world imagery, with 17,138 triplets spanning general, facial, and remote-sensing domains.

\item We develop a physics-motivated agentic data-construction workflow that converts shadow datasets with image-mask pairs into paired removal supervision through failure detection, conditional retry, candidate filtering and selection, structural diagnostics, and deterministic regional correction, making AgenticShadow readily extensible to new data sources.

\item We show that AgenticShadow supervision substantially improves cross-domain generalization across existing architectures and enables models with semantic and geometric priors---including a mask-free variant---to operate effectively under substantially broader real-world supervision.

\end{itemize}

\section{Related Work}

\noindent\textbf{Shadow Removal.}
Early methods removed shadows using illumination invariance~\citep{finlayson2002removing}, matting~\citep{wu2007natural}, texture consistency~\citep{liu2008texture}, and relationships between paired shadowed and lit regions~\citep{guo2013paired}. Learning-based approaches later formulated the task as image restoration through multi-context reasoning~\citep{qu2017deshadownet}, joint detection and removal~\citep{wang2018stacked}, shadow-image decomposition~\citep{le2019shadowdecomposition,le2022physics}, and hierarchical feature aggregation~\citep{cun2020ghostfree}. Recent methods adopt transformer architectures for global-context or homogenized restoration~\citep{guo2023shadowformer,xiao2024homoformer}, inpainting formulations~\citep{li2023inpainting}, diffusion models~\citep{guo2023shadowdiffusion,xu2025detail}, and semantic or geometric priors~\citep{lee2026phasr}. Although these methods improve performance on established benchmarks, their generalization across substantially different scenes and shadow configurations remains underexplored.

\noindent\textbf{Shadow Datasets and Benchmarks.}
Paired benchmarks including SRD~\citep{qu2017deshadownet} and ISTD~\citep{wang2018stacked} provide real shadow and shadow-free pairs but remain modest in scale and limited in their diversity of scenes, viewpoints, and shadow configurations. WSRD~\citep{vasluianu2023wsrd} adds high-resolution pairs but uses a controlled, close-range setup focused on object and surface interactions. Synthetic datasets increase scale through generated shadows~\citep{liu2021shadow} or fully rendered virtual scenes under direct and indirect illumination~\citep{xu2024omnisr}, but retain a domain gap with real images. Real-world datasets span general scenes in SBU~\citep{vicente2016largescale} and CUHK-Shadow~\citep{hu2021revisiting}, object-shadow associations in SOBA~\citep{wang2020instance}, documents~\citep{das2020intrinsic,li2023highresolution}, faces~\citep{liu2022blind,luo2026beyond}, videos in ViSha~\citep{chen2021triple}, and remote-sensing imagery in S-EO~\citep{masquil2025seo}. However, large detection-oriented collections typically provide only shadow images and masks because capturing shadow-free counterparts is usually impractical. AgenticShadow bridges this gap by combining broad real-world coverage with unified image-mask-target supervision.

\noindent\textbf{Generative Editing for Data Construction.}
Instruction-guided editors enable localized image modifications from natural-language requests~\citep{brooks2023instructpix2pix,zhang2023magicbrush,sheynin2024emuedit,zhao2024ultraedit}, but do not explicitly model shadow formation; their stochastic outputs may preserve shadows, alter scene content, or introduce physically inconsistent changes. Recent work couples image generation with model-based validation and correction for visual data construction~\citep{huang2026gennval,park2026seeandfix}. We adapt this general paradigm to shadow removal through physics-motivated candidate generation and evaluation, failure-driven retry, candidate selection, and deterministic regional correction.
\section{Agentic Dataset Construction}
\label{sec:construction}

\begin{figure}[!t]
    \centering
    \includegraphics[width=\linewidth]{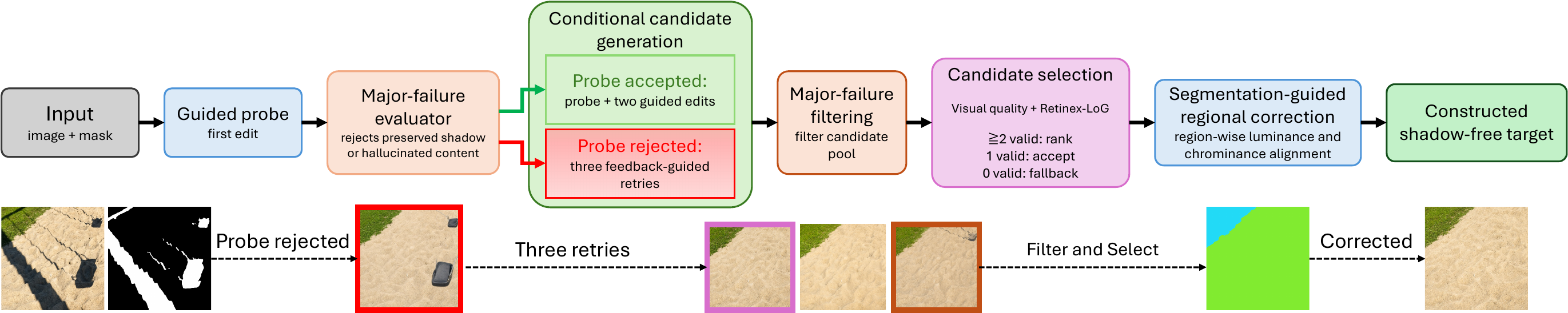}
    \caption{\textbf{Agentic dataset-construction pipeline.} A VLM evaluates an initial mask-guided probe, triggers additional guided edits or feedback-guided retries, filters major failures, and selects a candidate based on visual quality and Retinex-LoG diagnostics. Segmentation-guided regional correction produces the final shadow-free target. The bottom row illustrates a rejected-probe example.}
    \label{fig:pipeline}
\end{figure}

Given an image $I$ and shadow mask $M$, we construct a shadow-free target $\hat{I}$ while preserving scene geometry, surface detail, and non-shadow appearance. As summarized in \Figref{fig:pipeline}, GPT Image 2~\citep{openai2026models} produces candidate edits, while GPT-5 mini~\citep{singh2026openaigpt5card} serves as the vision-language model (VLM) that evaluates major failures, provides feedback for retry, and selects among valid candidates.
Detailed prompt designs and pipeline settings are provided in \Appref{sec:supp_construction_prompts}. Direct validation against color-corrected ISTD+ references is provided in \Appref{sec:supp_construction_istdplus}.

\paragraph{Shadow-formation grounding.}
A shadow represents reduced illumination caused by light occlusion, not a change in the underlying material or scene content. We encode this distinction in both the generation and evaluation prompts. The editor is asked to recover the appearance of the same surface under normal illumination, while the evaluator rejects candidates that reinterpret the shadow as an object, material boundary, or missing content. We compare this physics-motivated formulation with an otherwise identical pipeline using generic shadow removal prompts.

\paragraph{Probe-guided conditional generation.}
The pipeline begins with a mask-guided probe. A major-failure evaluator determines whether the probe retains the shadow by treating it as object-like content or hallucinates new scene content within the masked region. If the probe passes, we retain it and generate two additional guided candidates. If it fails, we discard it and generate three feedback-guided retry candidates from the original image and mask.

\paragraph{Candidate filtering and selection.}
Every candidate entering the selection pool must pass the major-failure evaluator. When multiple candidates remain, the VLM selector compares their shadow removal quality, scene fidelity, detail preservation, and artifact severity, supported by Retinex-LoG structural diagnostics. Because raw gradients are sensitive to illumination, we apply Retinex~\citep{jobson1997properties} in the log-luminance domain to suppress slowly varying illumination and use a Laplacian-of-Gaussian (LoG)~\citep{marr1980theory} operator to measure the remaining local structure. For an image $X$ with luminance $Y_X$, its Retinex-LoG response is

\begin{equation}
    F(X)=\left|\operatorname{LoG}\left(\log(Y_X+\epsilon)-G_{\sigma}*\log(Y_X+\epsilon)\right)\right|.
    \label{eq:retinex_log}
\end{equation}

Here, $G_{\sigma}$ estimates smooth illumination, and subtracting it from log-luminance produces a reflectance representation. Let $\Omega_M$ denote the valid pixels obtained by excluding a narrow band around the shadow boundary. Before comparison, both response maps are divided by the 95th-percentile response of the input over $\Omega_M$ and clipped to $[0,1]$, yielding $\widetilde{F}(I)$ and $\widetilde{F}(C)$. We then measure the structural discrepancy between candidate $C$ and input $I$ as

\begin{equation}
    D_{\mathrm{Retinex\text{-}LoG}}(C,I)
    =
    \operatorname{MAE}_{\Omega_M}
    \bigl(\widetilde{F}(C),\widetilde{F}(I)\bigr).
    \label{eq:retinex_log_distance}
\end{equation}

A lower discrepancy indicates better local structure preservation. We also measure the mean magnitudes of response decreases and increases to indicate lost and added detail. These diagnostics support, but do not replace, the selector's visual judgment. If one candidate passes, it is selected directly; if none pass, a fallback selector returns the least damaging candidate from the generated set.

\paragraph{Segmentation-guided regional correction.}
The selected proposal may remove the shadow successfully while exhibiting broad or regional color shifts. We segment the proposal into class-agnostic regions using CropFormer~\citep{qi2023high}. 
For each region, we compare the input and proposal only at reliable pixels outside a dilated shadow mask and estimate a constant offset from the channel-wise median of their YCbCr differences after outlier filtering. A local offset is applied when the region contains sufficient reliable pixels and improves low-frequency color agreement; otherwise, a global offset is used. Because this stage adjusts only brightness and chrominance, it does not explicitly copy textures, edges, or shadow structure from the input. The corrected image becomes the final target $\hat{I}$.

\section{AgenticShadow Benchmark}
\label{sec:benchmark}

\paragraph{Dataset composition.}
AgenticShadow combines four complementary sources of real shadow images: SBU~\citep{vicente2016largescale}, CUHK-Shadow~\citep{hu2021revisiting}, ASFW~\citep{luo2026beyond}, and S-EO~\citep{masquil2025seo}. We apply the agentic construction workflow described in \Secref{sec:construction} to construct a shadow-free target for each retained image-mask pair. As summarized in \Tabref{tab:dataset_composition}, the resulting benchmark contains 17,138 image-mask-target triplets, divided into 14,117 training and 3,021 test examples. SBU and CUHK-Shadow provide diverse general scenes, while ASFW and S-EO extend the benchmark to facial shadows and remote-sensing imagery. Source processing, cleaning, splits, and retained counts are detailed in \Appref{sec:supp_dataset_processing}.

\begin{table}[!t]
\centering
\caption{\textbf{Composition of AgenticShadow.} The four source domains provide complementary scene types and together form a benchmark of 17,138 image-mask-target triplets.}
\label{tab:dataset_composition}
\small
\renewcommand{\arraystretch}{1.15}
\begin{tabular*}{\linewidth}{@{\extracolsep{\fill}}lccccc@{}}
\toprule
& SBU & CUHK & ASFW & S-EO & \textbf{Total} \\
\midrule
Domain & General & General & Facial & Remote sensing & -- \\
Train & 3,974 & 8,335 & 928 & 880 & 14,117 \\
Test & 635 & 2,087 & 153 & 146 & 3,021 \\
\midrule
\textbf{Total} & \textbf{4,609} & \textbf{10,422} & \textbf{1,081} & \textbf{1,026} & \textbf{17,138} \\
\bottomrule
\end{tabular*}
\end{table}

\begin{figure}[!t]
\centering
\includegraphics[width=0.93\linewidth]{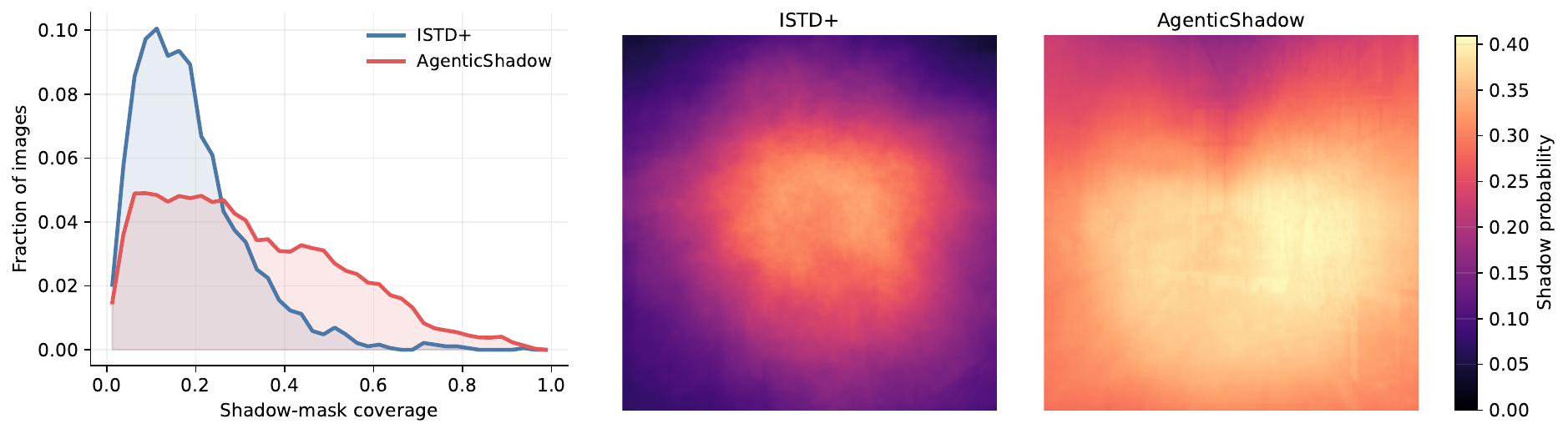}
\par\vspace{-0.3em}
\makebox[0.93\linewidth]{%
    \makebox[0.333\linewidth]{\small (a) Shadow coverage}%
    \makebox[0.667\linewidth]{\small (b) Shadow location}%
}

\par\vspace{0.5em}

\includegraphics[width=0.93\linewidth]{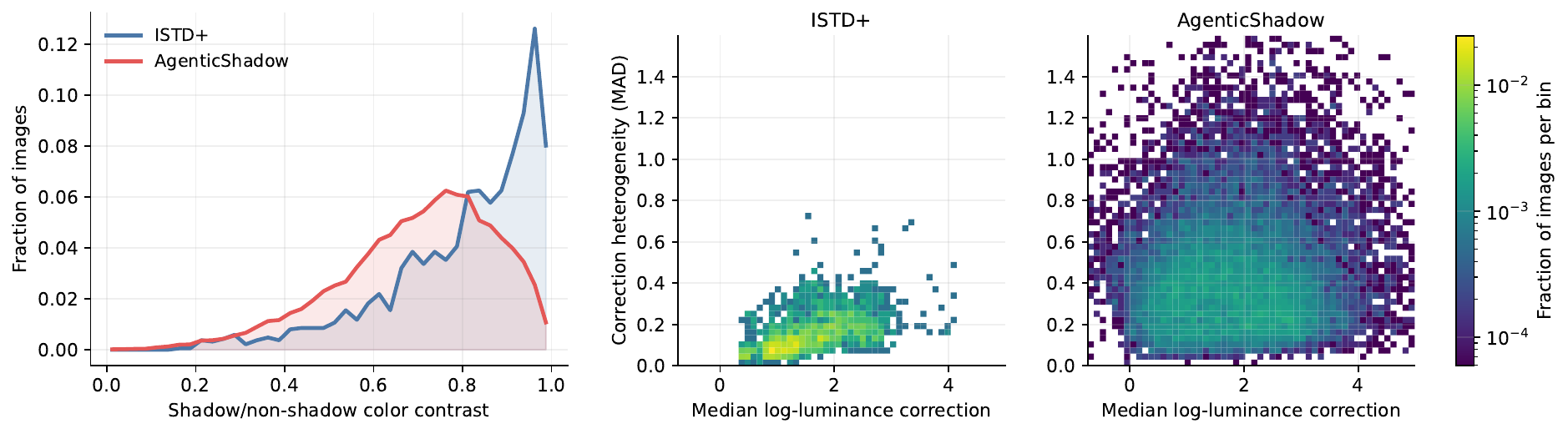}
\par\vspace{-0.3em}
\makebox[0.93\linewidth]{%
    \makebox[0.333\linewidth]{\small (c) Color contrast}%
    \makebox[0.667\linewidth]{\small (d) Correction diversity}%
}

\caption{\textbf{Dataset diversity compared with ISTD+.}
(a) Distribution of shadow-mask coverage.
(b) Average spatial distribution of shadow masks.
(c) Color contrast between shadow and non-shadow regions.
(d) Joint distribution of median log-luminance correction and within-shadow heterogeneity.
AgenticShadow contains larger and more spatially distributed shadows, more low-contrast cases, and substantially broader and more heterogeneous correction requirements.}
\label{fig:benchmark_statistics}
\end{figure}

\paragraph{Benchmark diversity.}
We compare AgenticShadow with ISTD+~\citep{le2019shadowdecomposition} using four complementary statistics computed over each complete benchmark. Shadow coverage is the fraction of pixels inside the shadow mask, while the location heatmap averages masks after resizing them to common image coordinates. Following the analysis of CUHK-Shadow~\citep{hu2021revisiting}, color contrast is measured by the $\chi^2$ distance between normalized RGB histograms of shadow and non-shadow pixels. To characterize correction diversity, we compute the signed log-luminance change from each shadow image to its target within the shadow region; its median measures the typical correction, while its median absolute deviation (MAD) measures spatial heterogeneity.

As shown in \Figref{fig:benchmark_statistics}, AgenticShadow contains substantially larger shadows, with median coverage of $28.6\%$ versus $16.1\%$ for ISTD+. Its heatmap shows broader shadow locations, while its lower median color contrast ($0.724$ versus $0.849$) indicates greater coverage of subtle, low-contrast shadows. Finally, its median log-luminance correction increases from $1.51$ to $1.83$, while correction heterogeneity increases from $0.141$ to $0.395$. These results demonstrate that AgenticShadow covers more diverse and spatially varying shadow removal requirements than ISTD+. 
Additional construction and semantic-diversity statistics are provided in \Appref{sec:supp_dataset_statistics}.

\paragraph{Evaluation protocol.}
We evaluate shadow removal methods separately on the SBU, CUHK-Shadow, ASFW, and S-EO test sets. Because their sizes differ substantially, we report per-domain results and a macro average that assigns equal weight to each domain. We report ISTD+ separately as an established benchmark. Reconstruction quality is measured using PSNR, SSIM, MAE, and LAB RMSE. 
Checkpoint, inference, and metric details are provided in \Appref{sec:supp_evaluation_protocol}.
\section{Experiments and Results}
\label{sec:experiments}

\noindent\textbf{Implementation Details.}
We implement the restoration experiments in PyTorch~\citep{paszke2019pytorch} and train on four NVIDIA RTX A6000 GPUs. Our PF baseline combines a pretrained HomoFormer~\citep{xiao2024homoformer} with frozen DINOv2-L/14~\citep{oquab2024dinov2} and Depth Anything V2-L~\citep{yang2024depthanythingv2} models, using input resolutions of 448 and 518 pixels, respectively. Unless otherwise stated, AgenticShadow-trained models use the natural concatenation of ISTD+ and AgenticShadow. PF is trained for 100 epochs with a total batch size of 16 using Adam~\citep{kingma2017adam}. The backbone and prior-module learning rates start at $10^{-4}$ and $3\times10^{-4}$ and decay to $10^{-6}$ under cosine scheduling. PF-MF follows the same configuration. Additional training details are provided in \Appref{sec:supp_pf_implementation}, with model complexity and runtime reported in \Appref{sec:supp_efficiency}.

\noindent\textbf{Evaluations.}
We evaluate agentic-construction quality on the 399-image ShadowRemovalRefine benchmark~\citep{hu2025shadowrefine} using Color Distribution Difference (CDD), which measures color discrepancy across annotated shadow boundaries; lower is better. We compare with SID~\citep{le2019shadowdecomposition}, ShadowFormer~\citep{guo2023shadowformer}, ShadowDiffusion~\citep{guo2023shadowdiffusion}, Inpaint4Shadow~\citep{li2023inpainting}, and ShadowRemovalRefine. Because generative editing is stochastic, we evaluate each agentic configuration over two independent full-set runs and average the results. For cross-domain evaluation, we test official ISTD+ checkpoints of SID, ShadowFormer, Inpaint4Shadow, StableSR~\citep{xu2025detail}, ShadowDiffusion, HomoFormer, and PhaSR~\citep{lee2026phasr} across all four AgenticShadow domains. We additionally compare ShadowDiffusion, HomoFormer, and PhaSR with counterparts trained on ISTD+ and AgenticShadow. We report per-domain and macro-average PSNR, SSIM, MAE, and LAB RMSE, with ISTD+ reported separately.

\subsection{Agentic Target-Construction Quality}
\label{sec:agentic_quality}

\begin{table*}[!t]
\centering
\footnotesize

\begin{minipage}[t]{0.60\linewidth}
\vspace{0pt}
\centering
\strut\textbf{(a) Quantitative comparison}\par
\vspace{0.35em}
\setlength{\tabcolsep}{4.5pt}
\renewcommand{\arraystretch}{1.44}
\begin{tabular}{lrrrr}
\toprule
Method & Mean & Std & Min & Max \\
\midrule
SID~\citep{le2019shadowdecomposition}
    & 38.00 & 45.05 & 1.66 & 353.64 \\
SF~\citep{guo2023shadowformer}
    & 33.63 & 52.74 & 0.41 & 427.34 \\
SD~\citep{guo2023shadowdiffusion}
    & 89.72 & 130.54 & 0.29 & 650.26 \\
I4S~\citep{li2023inpainting}
    & 31.57 & 48.39 & 0.43 & 402.44 \\
SRR~\citep{hu2025shadowrefine}
    & 14.06 & 26.08 & 0.24 & 215.40 \\
\midrule
\textbf{Ours-Physics}
    & \textbf{6.96}
    & \textbf{18.59}
    & \textbf{0.01}
    & \textbf{158.40} \\
\bottomrule
\end{tabular}
\end{minipage}
\hfill
\begin{minipage}[t]{0.38\linewidth}
\vspace{0pt}
\centering
\strut\textbf{(b) Pipeline ablation}\par
\vspace{0.35em}
\setlength{\tabcolsep}{4pt}
\renewcommand{\arraystretch}{1.12}
\begin{tabular}{llr}
\toprule
Prompt & Variant & CDD \\
\midrule
        & Probe                & 12.07 \\
Generic & Major-retry          & 11.77 \\
        & Visual selector      & 10.33 \\
\midrule
        & Probe                & 9.51 \\
        & Major-retry          & 7.78 \\
Physics & Visual selector      & 7.45 \\
        & Retinex-LoG          & 7.07 \\
        & Regional correction & \textbf{6.96} \\
\bottomrule
\end{tabular}
\end{minipage}

\caption{\textbf{Agentic target-construction quality on SRR~\citep{hu2025shadowrefine}.}
(a) Comparison with previous shadow removal methods.
(b) Ablation of the agentic construction pipeline.
CDD values are multiplied by $10^3$, agentic results are averaged over two
independent runs, and lower is better.}
\label{tab:agentic_quality}
\end{table*}

We assess both the final target quality and the contribution of each construction stage on the common normalized SRR benchmark described in \Appref{sec:supp_construction_normalization}.
As shown in \Tabref{tab:agentic_quality}(a), our final physics-motivated pipeline achieves a mean CDD of 0.006960, a 50.5\% reduction relative to the strongest previous method, SRR. It also yields the lowest standard deviation and lowest maximum CDD, indicating that the workflow improves average target quality while reducing severe construction failures.
See \Appref{sec:supp_construction_examples} for qualitative analyses and \Appref{sec:supp_construction_limitations} for limitations.

\Tabref{tab:agentic_quality}(b) isolates the contributions of the construction pipeline. Under generic prompting, major-failure retry and visual selection reduce mean CDD from 0.012073 to 0.010334. Shadow-formation grounding consistently outperforms its generic counterpart at the probe, retry, and selection stages, showing that explicitly treating shadows as illumination effects improves both candidate generation and evaluation. Retinex-LoG evidence and regional correction further reduce CDD to 0.006960. Together, these results validate both the agentic control flow and its physics-motivated grounding.

\subsection{Prior-Fusion Baseline}
\label{sec:pf_baseline}

\begin{figure}[!t]
    \centering
    \includegraphics[width=\linewidth]{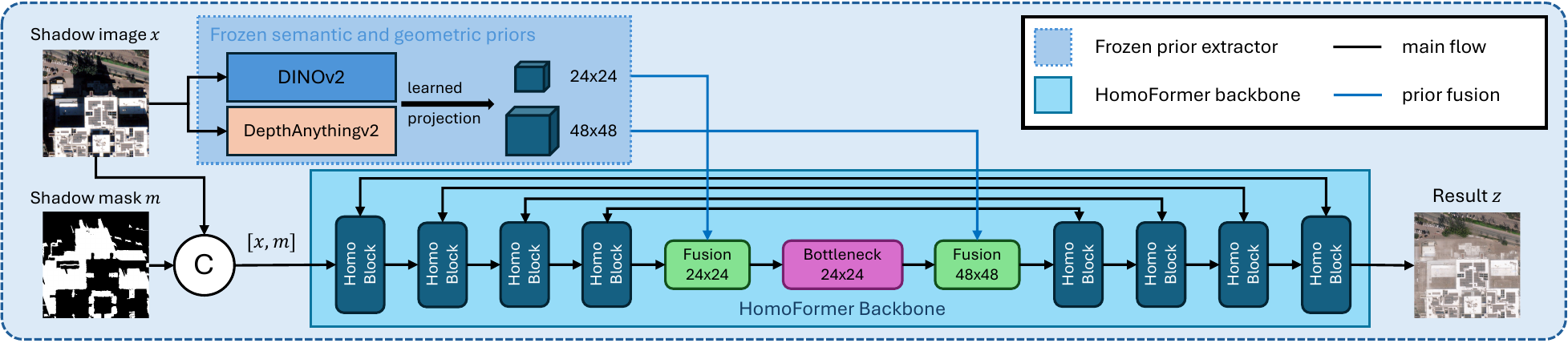}
    \caption{\textbf{Prior-Fusion (PF) baseline.} Frozen DINOv2~\citep{oquab2024dinov2} and Depth Anything V2~\citep{yang2024depthanythingv2} extract semantic and geometric priors from the shadow image. Learned projections align these priors with intermediate features, and zero-initialized residual fusion incorporates them at the $24\times24$ bottleneck input and the first $48\times48$ decoder stage.}
    \label{fig:method}
\end{figure}

\begin{table*}[!t]
\centering
\footnotesize

\begin{minipage}[t]{0.52\linewidth}
\vspace{0pt}
\centering
\strut\textbf{(a) Dataset--prior interaction}\par
\vspace{0.35em}
\setlength{\tabcolsep}{3.8pt}
\renewcommand{\arraystretch}{1.16}
\begin{tabular}{lcccc}
\toprule
& \multicolumn{2}{c}{ISTD+}
& \multicolumn{2}{c}{AgenticShadow} \\
\cmidrule(lr){2-3}\cmidrule(lr){4-5}
Model
& PSNR$\uparrow$ & LAB$\downarrow$
& PSNR$\uparrow$ & LAB$\downarrow$ \\
\midrule
HomoFormer$^{*}$
& 32.53 & 1.92 & 18.85 & 8.43 \\
PF$^{*}$
& 32.95 & 1.87 & 18.70 & 8.55 \\
\midrule
HomoFormer
& 32.92 & 1.85 & 21.95 & 5.87 \\
PF
& 32.74 & 1.86 & 22.50 & 5.55 \\
\bottomrule
\end{tabular}
\end{minipage}
\hfill
\begin{minipage}[t]{0.46\linewidth}
\vspace{0pt}
\centering
\strut\textbf{(b) Prior-component ablation}\par
\vspace{0.35em}
\setlength{\tabcolsep}{2.7pt}
\renewcommand{\arraystretch}{1.23}
\begin{tabular}{lcccc}
\toprule
& \multicolumn{4}{c}{AgenticShadow} \\
\cmidrule(lr){2-5}
Variant (E50)
& PSNR$\uparrow$ & SSIM$\uparrow$
& MAE$\downarrow$ & LAB$\downarrow$ \\
\midrule
HomoFormer
& 21.90 & 0.7161 & 15.36 & 5.912 \\
Depth only
& 22.13 & 0.7259 & 14.96 & 5.769 \\
Semantic only
& 22.46 & 0.7306 & 14.42 & 5.585 \\
PF
& \textbf{22.47} & \textbf{0.7312}
& \textbf{14.40} & \textbf{5.577} \\
\bottomrule
\end{tabular}
\end{minipage}

\caption{\textbf{Analysis of the Prior-Fusion baseline.}
(a) Interaction between training data and prior fusion, comparing ISTD+-only training with combined ISTD+ and AgenticShadow training; $^{*}$ denotes the former.
(b) Matched E50 prior ablation initialized from HomoFormer
E100~\citep{xiao2024homoformer}. AgenticShadow results are four-domain
macro averages.}
\label{tab:pf_analysis}
\end{table*}

To examine whether broader paired supervision enables more effective use of external priors, we construct a \textbf{Prior-Fusion (PF)} baseline based on HomoFormer~\citep{xiao2024homoformer}. Given a shadow image $x$ and mask $m$, PF predicts a shadow-free image in one feed-forward pass while incorporating frozen semantic and geometric features, as illustrated in \Figref{fig:method}. This deliberately simple design provides a controlled way to evaluate how training supervision affects the utility of external priors.

\paragraph{Prior extraction and fusion.}
We extract semantic features using DINOv2~\citep{oquab2024dinov2} and relative depth using Depth Anything V2~\citep{yang2024depthanythingv2}, keeping both models frozen. A learned $1\times1$ projection and spatial resizing align the semantic features, while a lightweight convolutional pyramid projects the depth map. Let $P_r^{\mathrm{sem}}$ and $P_r^{\mathrm{geo}}$ denote the resulting features at resolution $r\times r$. They are fused with HomoFormer tokens $H_r$ through
\begin{equation}
\widetilde{H}_r
=
H_r
+
\alpha_r\,\phi\!\left(P_r^{\mathrm{sem}}\right)
+
\beta_r\,\phi\!\left(P_r^{\mathrm{geo}}\right),
\qquad r\in\{24,48\},
\label{eq:prior_fusion}
\end{equation}
where $\phi(\cdot)$ converts feature maps into tokens, and $\alpha_r$ and $\beta_r$ are learned scalar strengths initialized to zero. Fusion is applied before the $24\times24$ bottleneck and at the first $48\times48$ decoder stage.

\paragraph{Training and variants.}
The prior extractors remain fixed, while the backbone, projections, and fusion strengths are optimized using the Charbonnier loss~\citep{charbonnier1994two}. We additionally evaluate semantic-only and depth-only variants. PF-MF removes the mask input and its reinjection at the final decoder and output stages while retaining the same prior-fusion design.

\Tabref{tab:pf_analysis}(a) shows that prior fusion alone is insufficient for generalization. Under ISTD+-only training, PF improves in-domain performance but not AgenticShadow results. With combined ISTD+ and AgenticShadow supervision, the same priors provide clear AgenticShadow gains while retaining comparable ISTD+ performance, indicating that broader supervision is needed to translate external priors into transferable improvements.
\Tabref{tab:pf_analysis}(b) provides a matched E50 component analysis initialized from HomoFormer E100. Both priors independently improve the backbone across all metrics. Their combination performs best overall: semantic information provides the larger gain, while relative depth contributes complementary geometric context.

\subsection{Cross-Domain Benchmark Evaluation}
\label{sec:cross_domain}

\begin{table}[!t]
\centering
\caption{\textbf{Cross-domain shadow removal evaluation.}
Official ISTD+ checkpoints are evaluated without adaptation, while $^\dagger$ denotes training with both ISTD+ and AgenticShadow. We report per-domain LAB RMSE and equal-weight macro averages across the four AgenticShadow domains. ISTD+ is reported separately. Best and second-best results are \textbf{bolded} and \underline{underlined}, respectively.}
\label{tab:cross_domain}
\small
\setlength{\tabcolsep}{3.5pt}
\renewcommand{\arraystretch}{1.08}
\resizebox{\linewidth}{!}{%
\begin{tabular}{lccccccccc}
\toprule
& \multicolumn{1}{c}{ISTD+}
& \multicolumn{4}{c}{AgenticShadow: LAB RMSE $\downarrow$}
& \multicolumn{4}{c}{AgenticShadow: Macro} \\
\cmidrule(lr){2-2}
\cmidrule(lr){3-6}
\cmidrule(lr){7-10}
Method
& LAB $\downarrow$
& SBU
& CUHK
& ASFW
& S-EO
& PSNR $\uparrow$
& SSIM $\uparrow$
& MAE $\downarrow$
& LAB $\downarrow$ \\
\midrule

\multicolumn{10}{l}{\textit{Official ISTD+ checkpoints}} \\
SID~\citep{le2019shadowdecomposition}
& 2.89 & 7.25 & 10.86 & 7.48 & 10.23
& 18.28 & 0.6404 & 23.77 & 8.96 \\
ShadowFormer~\citep{guo2023shadowformer}
& 1.95 & 6.63 & 10.01 & 7.35 & 10.74
& 18.57 & 0.6592 & 23.36 & 8.68 \\
Inpaint4Shadow~\citep{li2023inpainting}
& 2.77 & 7.12 & 9.72 & 7.58 & 10.08
& 18.59 & 0.6119 & 23.06 & 8.63 \\
StableSR~\citep{xu2025detail}
& 2.03 & 7.32 & 8.84 & 10.01 & 10.93
& 17.90 & 0.6506 & 25.23 & 9.28 \\
ShadowDiffusion~\citep{guo2023shadowdiffusion}
& 1.94 & 6.83 & 9.09 & 7.47 & 10.50
& 18.72 & 0.6563 & 22.67 & 8.47 \\
PhaSR~\citep{lee2026phasr}
& 2.13 & 7.32 & 8.71 & 10.36 & 10.92
& 17.90 & 0.6556 & 24.75 & 9.33 \\
HomoFormer~\citep{xiao2024homoformer}
& 1.92 & 6.78 & 9.08 & 7.00 & 10.87
& 18.85 & 0.6597 & 22.36 & 8.43 \\

\midrule
\multicolumn{10}{l}{\textit{With AgenticShadow supervision}} \\
ShadowDiffusion$^\dagger$
& 2.06 & 5.84 & 7.40 & 5.97 & 7.98
& 20.47 & 0.6952 & 18.58 & 6.80 \\
PhaSR$^\dagger$
& 2.11 & 4.92 & 6.19 & 5.31 & 6.90
& 22.06 & \underline{0.7309} & 15.44 & 5.83 \\
HomoFormer$^\dagger$
& \textbf{1.85} & 4.90 & 6.23 & 5.33 & 7.02
& 21.95 & 0.7181 & 15.25 & 5.87 \\
\textbf{PF}
& \underline{1.86}
& \textbf{4.73}
& \textbf{6.05}
& \textbf{4.67}
& \textbf{6.76}
& \textbf{22.50}
& \textbf{0.7314}
& \textbf{14.36}
& \textbf{5.55} \\
PF-MF
& 1.95
& \underline{4.75}
& \underline{6.11}
& \underline{4.97}
& \underline{6.78}
& \underline{22.36}
& 0.7303
& \underline{14.67}
& \underline{5.65} \\

\bottomrule
\end{tabular}%
}
\end{table}

\begin{figure}[!t]
\centering
\includegraphics[width=\linewidth]{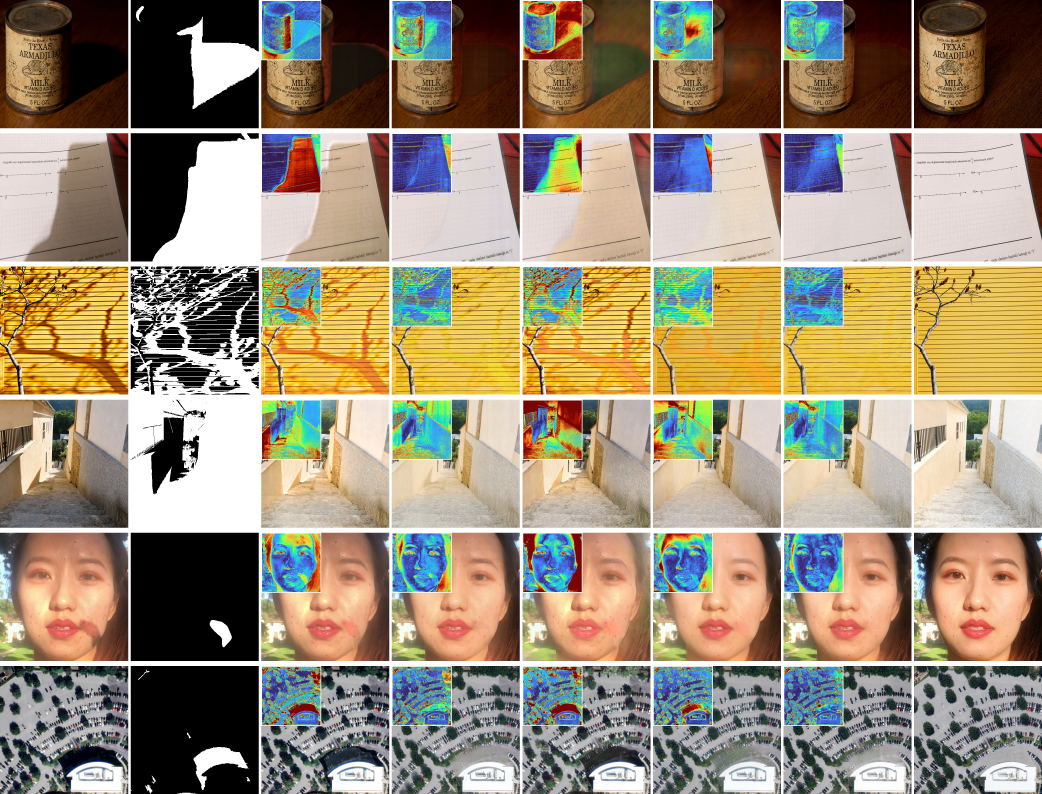}
\makebox[\linewidth]{%
\makebox[0.125\linewidth]{\scriptsize Input}%
\makebox[0.125\linewidth]{\scriptsize Mask}%
\makebox[0.125\linewidth]{\scriptsize HomoFormer}%
\makebox[0.125\linewidth]{\scriptsize HomoFormer$^\dagger$}%
\makebox[0.125\linewidth]{\scriptsize PhaSR}%
\makebox[0.125\linewidth]{\scriptsize PhaSR$^\dagger$}%
\makebox[0.125\linewidth]{\scriptsize PF}%
\makebox[0.125\linewidth]{\scriptsize Target}%
}
\caption{\textbf{Qualitative cross-domain comparison.}
We compare HomoFormer~\citep{xiao2024homoformer} and PhaSR~\citep{lee2026phasr}, their counterparts trained with AgenticShadow ($^\dagger$), and PF. Models pretrained on ISTD+~\citep{le2019shadowdecomposition} often leave residual shadows or color shifts on diverse AgenticShadow scenes, while training with AgenticShadow substantially improves removal. PF consistently produces outputs closer to the targets than the AgenticShadow-trained counterparts. Colored insets show mean absolute RGB error using a shared scale.}
\label{fig:cross_domain_qualitative}
\end{figure}

We first evaluate official ISTD+ checkpoints on the four AgenticShadow domains without adaptation, exposing their cross-domain limitations. We then retrain ShadowDiffusion, HomoFormer, and PhaSR using ISTD+ and AgenticShadow to measure the effect of broader supervision. PF follows the same combined-data setting as its HomoFormer backbone.

As shown in \Tabref{tab:cross_domain}, official checkpoints obtain macro LAB RMSE values of 8.43--9.33. AgenticShadow supervision reduces this range to 5.83--6.80 for the three retrained architectures, demonstrating a substantial generalization benefit from the additional data. PF performs best across every AgenticShadow domain and all macro metrics while remaining competitive on ISTD+. Complete per-domain and cross-dataset results appear in \Appref{sec:supp_per_domain} and \Appref{sec:supp_cross_dataset}.

\Figref{fig:cross_domain_qualitative} provides a visual counterpart to these results. Official ISTD+ checkpoints frequently retain shadows or introduce color shifts on unfamiliar scenes, while AgenticShadow-trained counterparts produce substantially cleaner outputs across general, facial, and remote-sensing imagery. PF further improves target agreement in the selected cases, consistent with the quantitative evaluation.

\subsection{Mask-Free and Generalization}
\label{sec:mask_free}

\begin{figure}[!t]
\centering
\includegraphics[width=\linewidth]{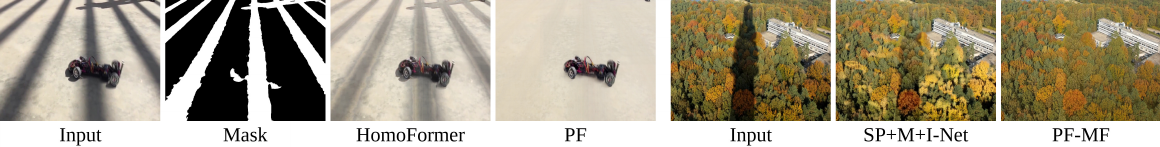}

\makebox[0.56\linewidth]{\small (a) PF on ViSha}%
\hfill
\makebox[0.42\linewidth]{\small (b) PF-MF on SBU-TimeLapse}%

\caption{\textbf{Generalization to shadow videos.}
(a) PF is compared with pretrained HomoFormer~\citep{xiao2024homoformer} on ViSha~\citep{chen2021triple} using the provided masks.
(b) PF-MF is compared with SP+M+I-Net on SBU-TimeLapse~\citep{le2022physics} without input masks.
In both settings, PF and PF-MF produce cleaner and more complete shadow removal than the compared methods.}
\label{fig:video_generalization}
\end{figure}

We evaluate PF-MF to assess whether AgenticShadow supervision and external priors reduce reliance on annotated masks. As shown in \Tabref{tab:cross_domain}, it closely matches PF, trailing by only 0.14\,dB PSNR and 0.10 LAB RMSE. \Figref{fig:video_generalization} shows video generalization: PF produces cleaner removal than pretrained HomoFormer on ViSha~\citep{chen2021triple} using masks, while PF-MF operates directly on SBU-TimeLapse~\citep{le2022physics} frames without masks and outperforms SP+M+I-Net. These results support video use where frame-level shadow masks are unavailable; complete videos and further qualitative comparisons are provided in \Appref{sec:supp_qualitative}.

\section{Conclusion}

We introduced AgenticShadow, a benchmark of 17,138 image-mask-target triplets constructed from diverse real shadow images through a physics-motivated agentic workflow. Our evaluation validates the constructed targets, shows that AgenticShadow offers substantially greater cross-domain diversity than existing paired benchmarks, and demonstrates that its supervision consistently improves representative shadow removal architectures. We also study a simple Prior-Fusion baseline, which achieves the strongest overall performance across the four AgenticShadow domains; its mask-free variant remains competitive and extends shadow removal to settings without annotated masks. Together, these results establish AgenticShadow as a resource for robust real-world shadow removal.

\subsection*{AI Use Statement}

Generative AI is part of our dataset-construction workflow. GPT Image 2 performs image normalization and generates candidate targets, while GPT-5 mini evaluates failures, guides retries, and selects candidates. Generative AI also assisted with pipeline implementation, experimental analysis, preparation of supporting artifacts, and manuscript editing. The authors reviewed all AI-assisted outputs, verified the reported results, and checked manuscript edits for accuracy. The authors take full responsibility for the final content of this work.

\bibliography{iclr2027_conference}
\bibliographystyle{iclr2027_conference}

\appendix
\section{AgenticShadow Dataset Details}
\label{sec:supp_dataset}

\subsection{Source Processing, Cleaning, and Splits}
\label{sec:supp_dataset_processing}

AgenticShadow is constructed from four existing shadow-image sources: SBU~\citep{vicente2016largescale}, CUHK-Shadow~\citep{hu2021revisiting}, ASFW~\citep{luo2026beyond}, and S-EO~\citep{masquil2025seo}. Because these sources differ in organization, annotation format, and image quality, we process each separately before target construction.

For SBU, we pair the original images with the refined training masks and relabeled test masks released by SILT~\citep{yang2023silt}. For CUHK-Shadow, we retain only pairs listed in the official training, validation, and test manifests. Duplicate review excludes 102 SBU and 70 CUHK-Shadow examples.

To place the heterogeneous SBU and CUHK-Shadow images on a common resolution and quality level, we apply the image-normalization procedure described in \Appref{sec:supp_construction_normalization}. Seven SBU images are excluded because normalization is blocked by content moderation. ASFW already provides aligned $512\times512$ image-mask pairs and therefore requires no normalization.

The S-EO release provides RGB images and Min-DSM-derived shadow annotations in separate archives. We match them by AOI and view identifiers, obtaining 19,162 pairs from 20,275 RGB images and 19,163 complete mask groups. Because many S-EO views contain large empty or invalid zero-valued regions within the AOI, we prioritize candidate crops with the highest valid-image coverage for visual review, aided by the released uncertainty, unseen-pixel, and vegetation annotations. We visually inspect 1,119 candidate crops and retain 1,026 after rejecting candidates with poor image or shadow mask quality and removing overlapping selections.

As summarized in \Tabref{tab:dataset_processing}, we produce 17,151 usable image-mask pairs before target generation. Candidate generation excludes five SBU and five CUHK-Shadow examples, while regional correction excludes three additional CUHK-Shadow examples, yielding the final 17,138 triplets.

SBU and ASFW retain their source splits. For CUHK-Shadow, the official validation set is assigned to training, while the test split is preserved. S-EO uses an AOI-disjoint split with 152 training and 25 test AOIs, ensuring that all crops and source views from each AOI remain in the same split. Duplicate review is completed before split finalization to prevent train-test overlap.

\begin{table*}[!t]
\centering
\caption{\textbf{AgenticShadow processing and retention.}
Eligible denotes matched source image-mask pairs, selected denotes examples retained after duplicate review or crop selection, usable denotes inputs ready for target construction after source-specific preprocessing, and generated denotes successful target generation before regional correction.}
\label{tab:dataset_processing}
\small
\setlength{\tabcolsep}{7pt}
\renewcommand{\arraystretch}{1.1}
\begin{tabular*}{\textwidth}{@{\extracolsep{\fill}}lrrrrrrr@{}}
\toprule
& & & & & \multicolumn{3}{c}{Final triplets} \\
\cmidrule(lr){6-8}
Source & Eligible & Selected & Usable & Generated & Train & Test & Total \\
\midrule
SBU  & 4,723  & 4,621  & 4,614  & 4,609  & 3,974 & 635   & 4,609 \\
CUHK & 10,500 & 10,430 & 10,430 & 10,425 & 8,335 & 2,087 & 10,422 \\
ASFW & 1,081  & 1,081  & 1,081  & 1,081  & 928   & 153   & 1,081 \\
S-EO & 19,162 & 1,026  & 1,026  & 1,026  & 880   & 146   & 1,026 \\
\midrule
\textbf{Total}
& \textbf{35,466} & \textbf{17,158} & \textbf{17,151}
& \textbf{17,141} & \textbf{14,117} & \textbf{3,021}
& \textbf{17,138} \\
\bottomrule
\end{tabular*}
\end{table*}

\subsection{Construction and Diversity Statistics}
\label{sec:supp_dataset_statistics}

\paragraph{Construction outcomes.}
We retain the generation and evaluation decisions for every constructed case. As shown in \Tabref{tab:construction_outcomes}, the initial probe passes the major-failure evaluator for 97.93\% of generated targets, while 2.07\% enter the feedback-guided retry branch. After candidate filtering, 99.52\% of cases contain at least two valid candidates for Retinex-LoG-supported selection. Only 68 cases contain a single passing candidate, and 15 cases use the least-damaging fallback because no candidate passes the major-failure evaluator. For the largest source, CUHK-Shadow, we track \$822.06 in API cost for generating 10,425 targets, including retries, averaging approximately \$0.079 per target. Subsequent segmentation, correction, and auditing are performed locally.

\begin{figure*}[!t]
\centering
\includegraphics[width=\textwidth]{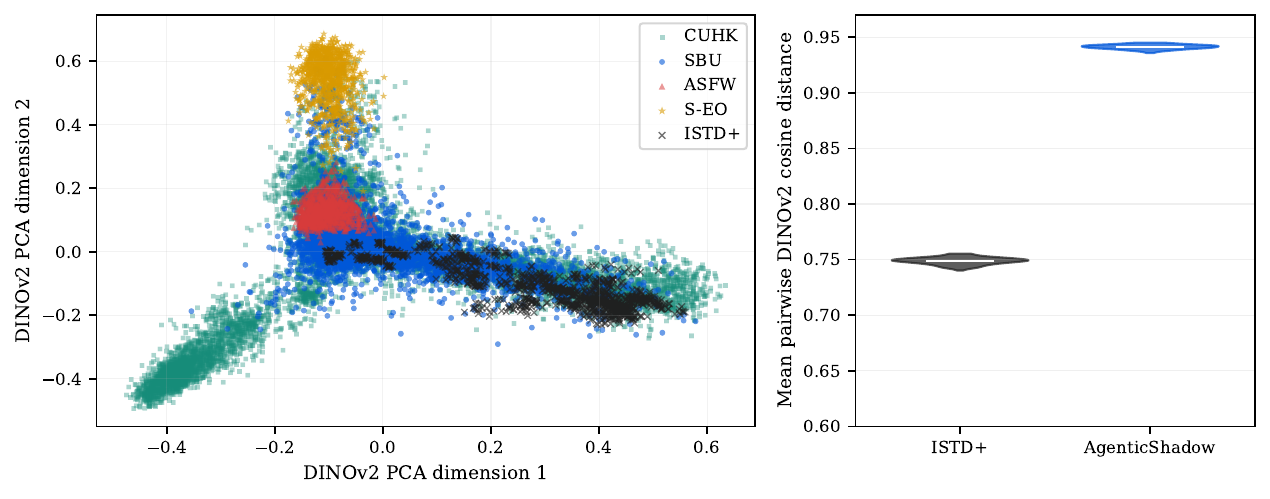}
\makebox[\textwidth]{%
    \makebox[0.63\textwidth]{\small (a) DINOv2 embedding space}%
    \makebox[0.37\textwidth]{\small (b) Size-controlled diversity}%
}
\caption{\textbf{Semantic scene diversity compared with ISTD+.}
(a) Joint PCA visualization of DINOv2-L/14 embeddings for all training images.
(b) Mean pairwise DINOv2 cosine distance over 100 size-matched resamples of 1,000 images; higher values indicate greater semantic diversity. Distances are computed in the original embedding space rather than the two-dimensional projection.}
\label{fig:scene_diversity}
\end{figure*}

\paragraph{Semantic scene diversity.}
We compare the semantic diversity of AgenticShadow and ISTD+~\citep{le2019shadowdecomposition} using normalized CLS embeddings from frozen DINOv2-L/14~\citep{oquab2024dinov2}. \Figref{fig:scene_diversity}(a) jointly projects all 14,117 AgenticShadow and 1,330 ISTD+ training images using PCA. AgenticShadow occupies a visibly broader embedding space, with facial and remote-sensing imagery extending into distinct semantic regions.
Because visual coverage grows with dataset size, we use a size-controlled comparison in \Figref{fig:scene_diversity}(b). We sample 1,000 images without replacement from each dataset and compute the mean pairwise cosine distance in the original DINOv2 embedding space, repeating this procedure 100 times. A larger distance means that two randomly selected images are less semantically similar and therefore indicates greater scene diversity. AgenticShadow obtains $0.9413\pm0.0020$, compared with $0.7487\pm0.0032$ for ISTD+, where values denote the mean and standard deviation across 100 resamples, corresponding to a 25.7\% increase.

\begin{table*}[!t]
\centering
\caption{\textbf{Dataset-wide target-construction outcomes.}
Pass and retry describe the initial routing decision. Multi, single, and fallback are mutually exclusive selection outcomes. Percentages are computed within each source and overall.}
\label{tab:construction_outcomes}
\small
\setlength{\tabcolsep}{5pt}
\renewcommand{\arraystretch}{1.05}
\begin{tabular*}{\textwidth}{@{\extracolsep{\fill}}lrrrrrr@{}}
\toprule
& & \multicolumn{2}{c}{Probe routing}
& \multicolumn{3}{c}{Final selection} \\
\cmidrule(lr){3-4}\cmidrule(lr){5-7}
Source & Generated & Pass & Retry & Multi & Single & Fallback \\
\midrule

\multirow[c]{2}{*}{SBU}
& \multirow[c]{2}{*}{4,609}
& 4,500 & 109 & 4,588 & 19 & 2 \\
& & {\scriptsize (97.64\%)} & {\scriptsize (2.36\%)}
& {\scriptsize (99.54\%)} & {\scriptsize (0.41\%)}
& {\scriptsize (0.04\%)} \\
\midrule

\multirow[c]{2}{*}{CUHK}
& \multirow[c]{2}{*}{10,425}
& 10,193 & 232 & 10,364 & 48 & 13 \\
& & {\scriptsize (97.77\%)} & {\scriptsize (2.23\%)}
& {\scriptsize (99.41\%)} & {\scriptsize (0.46\%)}
& {\scriptsize (0.12\%)} \\
\midrule

\multirow[c]{2}{*}{ASFW}
& \multirow[c]{2}{*}{1,081}
& 1,081 & 0 & 1,081 & 0 & 0 \\
& & {\scriptsize (100\%)} & {\scriptsize (0\%)}
& {\scriptsize (100\%)} & {\scriptsize (0\%)}
& {\scriptsize (0\%)} \\
\midrule

\multirow[c]{2}{*}{S-EO}
& \multirow[c]{2}{*}{1,026}
& 1,013 & 13 & 1,025 & 1 & 0 \\
& & {\scriptsize (98.73\%)} & {\scriptsize (1.27\%)}
& {\scriptsize (99.90\%)} & {\scriptsize (0.10\%)}
& {\scriptsize (0\%)} \\
\midrule

\multirow[c]{2}{*}{\textbf{Total}}
& \multirow[c]{2}{*}{\textbf{17,141}}
& \textbf{16,787} & \textbf{354} & \textbf{17,058}
& \textbf{68} & \textbf{15} \\
& & {\scriptsize\textbf{(97.93\%)}}
& {\scriptsize\textbf{(2.07\%)}}
& {\scriptsize\textbf{(99.52\%)}}
& {\scriptsize\textbf{(0.40\%)}}
& {\scriptsize\textbf{(0.09\%)}} \\
\bottomrule
\end{tabular*}
\end{table*}

\section{Agentic Construction Details}
\label{sec:supp_construction}

\subsection{Prompt Design and Pipeline Settings}
\label{sec:supp_construction_prompts}

The pipeline uses GPT Image 2~\citep{openai2026models} for candidate generation and GPT-5 mini~\citep{singh2026openaigpt5card} for evaluation and selection. Each case starts with one mask-guided probe. Passing probes are retained and followed by two additional candidates, for three editing calls total; rejected probes are replaced by three retries using evaluator feedback, for four calls total. Every candidate undergoes the same major-failure evaluation. A single passing candidate is selected directly, multiple passing candidates are compared using visual assessment supported by Retinex-LoG diagnostics, and if none pass, the least-damaging fallback selector is used.

\begin{figure*}[!t]
\centering
\includegraphics[width=\textwidth]{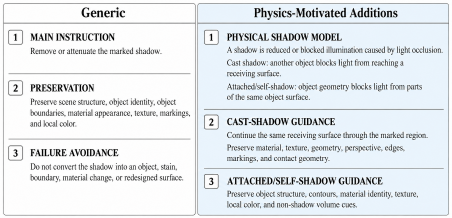}
\caption{\textbf{Generic prompting and physics-motivated additions.}
Both settings use identical inputs, preservation requirements, evaluator schemas, and agentic control flow. Physics-motivated prompting additionally defines cast and attached shadows as illumination effects caused by light occlusion and provides type-specific guidance for preserving the underlying surface or object.}
\label{fig:supp_prompt_design}
\end{figure*}

The two prompt families differ only in their shadow-formation grounding, as summarized in \Figref{fig:supp_prompt_design}. The generic prompts request shadow removal while preserving scene structure, object identity, materials, texture, markings, and local color. The physics-motivated prompts retain these requirements and additionally distinguish cast shadows on receiving surfaces from attached or self-shadows on the same object. This grounding is applied consistently to generation, retry, evaluation, and selection. The evaluator schema remains unchanged and rejects only two major failures: preserving a substantial shadow as coherent scene content or hallucinating semantic content inside the mask. This controlled design isolates the effect of shadow-formation grounding from the agentic control flow.

\subsection{Validation against Captured ISTD+ References}
\label{sec:supp_construction_istdplus}

\begin{figure*}[!t]
\centering
\includegraphics[width=\textwidth]{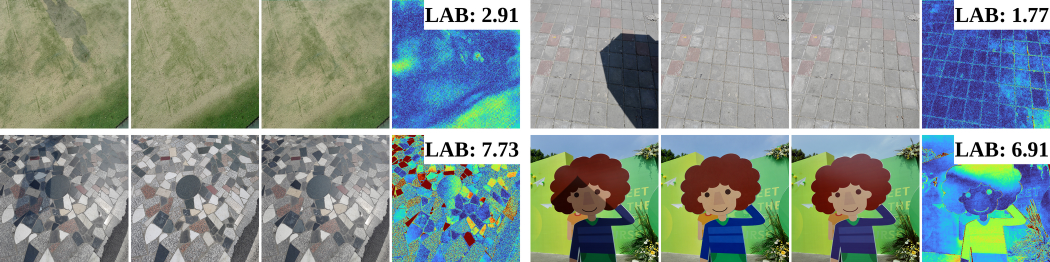}

\begin{minipage}[t]{0.495\textwidth}
\centering
\makebox[\linewidth][c]{%
    \makebox[0.25\linewidth][c]{\small Input}%
    \makebox[0.25\linewidth][c]{\small Constructed}%
    \makebox[0.25\linewidth][c]{\small Reference}%
    \makebox[0.25\linewidth][c]{\small Error}%
}
\end{minipage}%
\hfill
\begin{minipage}[t]{0.495\textwidth}
\centering
\makebox[\linewidth][c]{%
    \makebox[0.25\linewidth][c]{\small Input}%
    \makebox[0.25\linewidth][c]{\small Constructed}%
    \makebox[0.25\linewidth][c]{\small Reference}%
    \makebox[0.25\linewidth][c]{\small Error}%
}
\end{minipage}

\caption{\textbf{Construction against captured ISTD+ references.}
Each group shows the shadow input, our constructed target, the color-corrected ISTD+ reference, and the mean absolute RGB error map using a shared 0--50 scale. The top row presents two low-error cases, while the bottom row presents two higher-error cases that expose remaining texture, detail, and color discrepancies.}
\label{fig:supp_istdplus_construction}
\end{figure*}

We first test whether the construction workflow can recover known paired targets by applying it to all 540 ISTD+ test images and masks~\citep{le2019shadowdecomposition}. The workflow receives only the shadow image and mask, uses no ISTD+ reference during construction, and receives no task-specific training on ISTD+. We apply the same candidate generation, evaluation, Retinex-LoG-supported selection, and segmentation-guided regional correction used to construct AgenticShadow. Against the color-corrected ISTD+ references, the resulting targets achieve 25.11 PSNR, 0.7380 SSIM, 10.95 MAE, and 3.85 LAB RMSE. This direct comparison shows that the workflow can approximate the color-corrected ISTD+ references without using them during construction. Representative cases are shown in \Figref{fig:supp_istdplus_construction}. The top row presents low-error cases with shallow and stronger shadows, showing close agreement with the references across different shadow intensities. The bottom row presents higher-error cases with larger differences in local color and fine appearance. Nevertheless, the dominant shadows are removed while the overall scene content remains consistent.

\subsection{Image Normalization and API Usage}
\label{sec:supp_construction_normalization}

Because the source images vary substantially in resolution and image quality, we use a preprocessing-only GPT Image 2 pass to produce standardized $512\times512$ images while preserving the scene and original shadow appearance. The normalization prompt explicitly prohibits shadow removal, relighting, reframing, scene modification, or object addition and removal. For the ShadowRemovalRefine (SRR) benchmark~\citep{hu2025shadowrefine}, we apply this procedure to each image and spatially align its mask to the normalized output.

\begin{figure*}[!t]
\centering
\includegraphics[width=\textwidth]{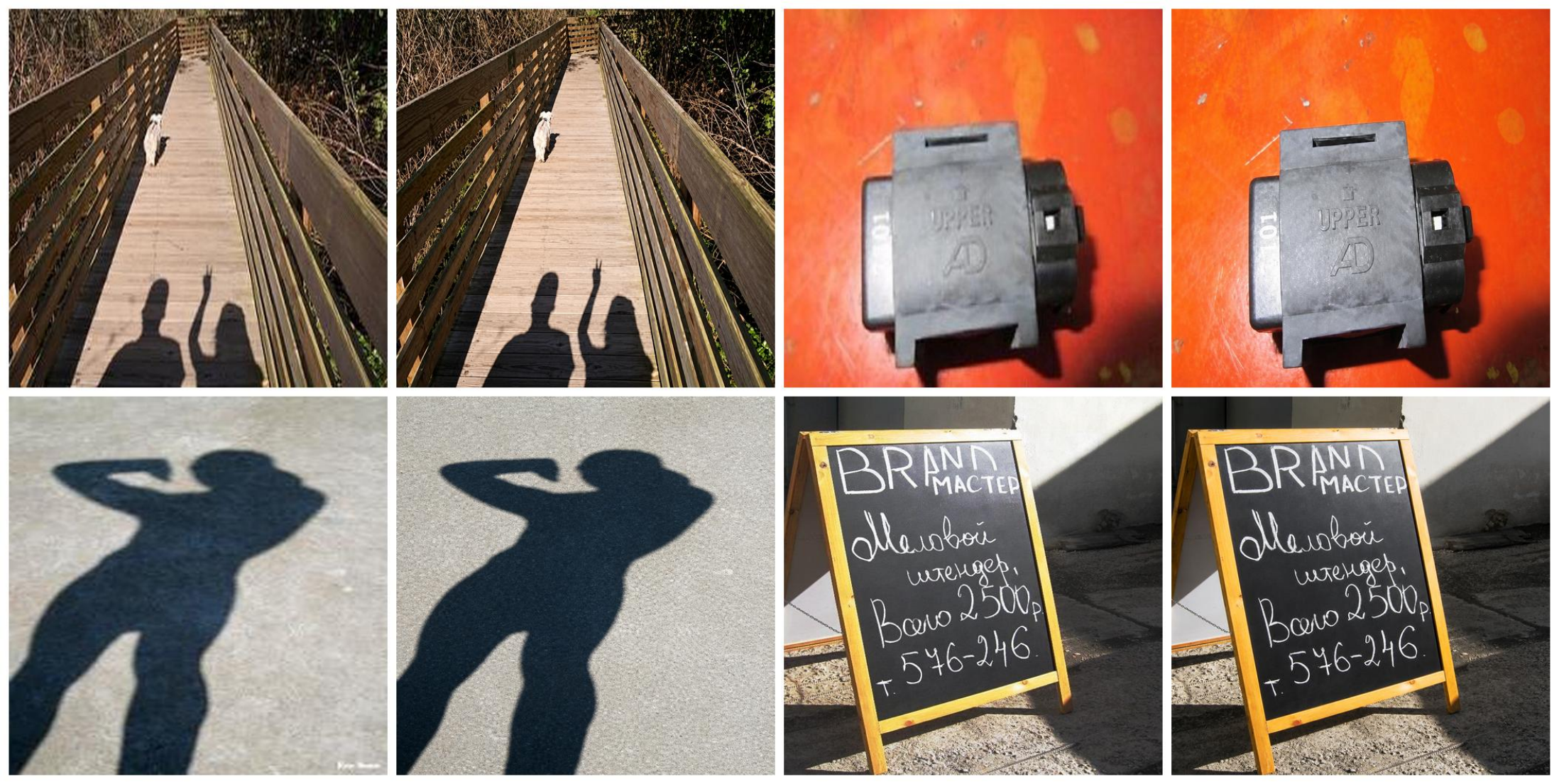}
\makebox[\textwidth]{%
    \makebox[0.25\textwidth]{\small Original}%
    \makebox[0.25\textwidth]{\small Normalized $512\times512$}%
    \makebox[0.25\textwidth]{\small Original}%
    \makebox[0.25\textwidth]{\small Normalized $512\times512$}%
}
\caption{\textbf{Image normalization examples.}
The first three pairs show low-resolution inputs standardized to $512\times512$, while the last pair shows a higher-resolution input normalized to the same resolution. The preprocessing pass improves resolution and image quality while preserving the scene and original shadow appearance.}
\label{fig:supp_image_normalization}
\end{figure*}

SRR originally contains 400 images with diverse resolutions and aspect ratios. One image is excluded after an API moderation block, leaving 399 evaluation cases. As illustrated in \Figref{fig:supp_image_normalization}, each image is normalized before candidate generation. All compared methods and agentic configurations receive the same normalized images and aligned masks, ensuring consistent quantitative comparisons.

\begin{table*}[!t]
\centering
\caption{\textbf{API usage for complete SRR runs.}
Each run processes all 399 evaluation cases. Pass and retry indicate the routing decision for the initial probe. Edits include the probe and subsequent guided or retry candidates, and each edit receives one major-failure evaluation. Selector counts exclude cases with exactly one passing candidate, which are selected directly.}
\label{tab:supp_api_usage}
\small
\setlength{\tabcolsep}{7pt}
\renewcommand{\arraystretch}{1.1}
\begin{tabular*}{\textwidth}{@{\extracolsep{\fill}}lcccccc@{}}
\toprule
Prompt setting & Cases & Pass & Retry & Edits & Evaluations & Selectors \\
\midrule
Generic
& 399 & 391 & 8 & 1,205 & 1,205 & 394 \\
Physics-motivated
& 399 & 389 & 10 & 1,207 & 1,207 & 397 \\
\bottomrule
\end{tabular*}
\end{table*}

\Tabref{tab:supp_api_usage} summarizes the API calls required by complete generic and physics-motivated runs. Their similar usage reflects the identical control flow, with shadow-formation grounding as the principal difference. The agentic workflow is intended for offline target construction rather than direct deployment: a complete 399-image run takes approximately 14 hours and incurs an estimated API cost of about \$29, whereas specialized learned models operate on the order of one second per image. General-purpose generative editors also remain stochastic and can misinterpret shadow structure, motivating the evaluation, filtering, and correction stages. The resulting targets can instead be constructed once and used to train efficient specialized models for deployment.

\subsection{Additional Qualitative Examples}
\label{sec:supp_construction_examples}

\begin{figure*}[!t]
\centering
\includegraphics[width=0.5\textwidth]{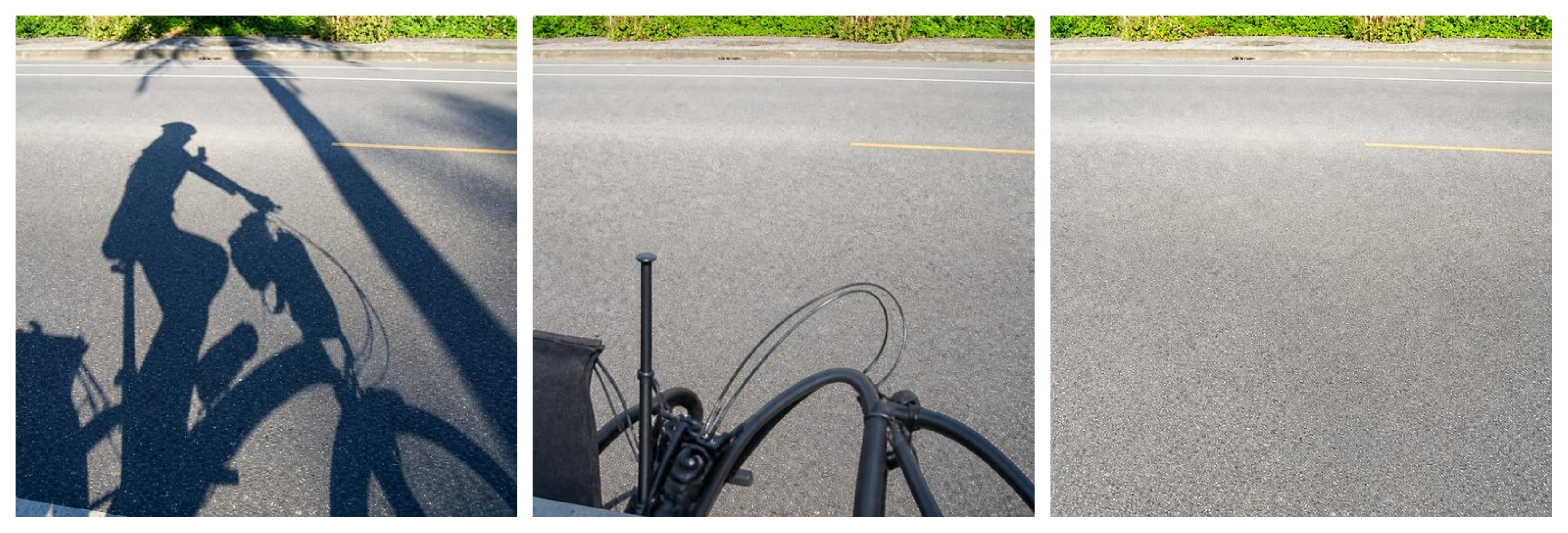}%
\includegraphics[width=0.5\textwidth]{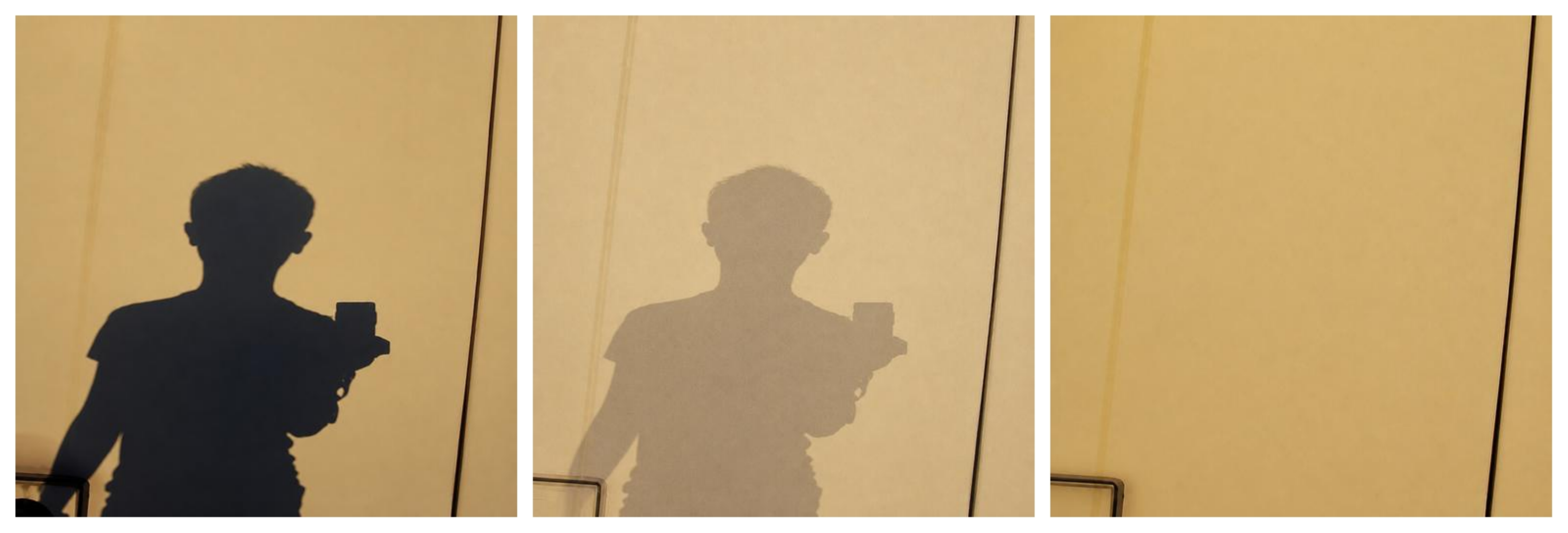}
\makebox[\textwidth]{%
    \makebox[0.166\textwidth]{\small Input}%
    \makebox[0.166\textwidth]{\small Probe}%
    \makebox[0.166\textwidth]{\small Retry}%
    \makebox[0.166\textwidth]{\small Input}%
    \makebox[0.166\textwidth]{\small Probe}%
    \makebox[0.166\textwidth]{\small Retry}%
}
\caption{\textbf{Effect of major-failure retry.}
The evaluator rejects probes with severe residual shadow structure or object-like hallucination in the masked region. Feedback-guided retry produces cleaner candidates before final selection.}
\label{fig:supp_major_retry}
\end{figure*}

\paragraph{Major-failure retry.}
\Figref{fig:supp_major_retry} shows cases in which the initial probe preserves the shadow as coherent scene content or introduces an object-like interpretation of the masked region. The major-failure evaluator rejects these probes and supplies feedback for retry generation, producing candidates with more complete shadow removal and better scene preservation.

\begin{figure*}[!t]
\centering
\includegraphics[width=\textwidth]{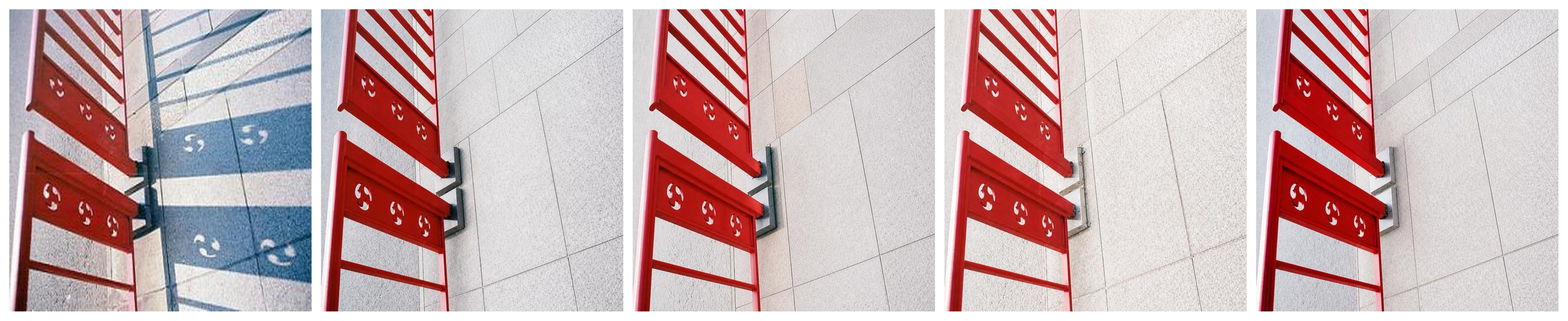}
\includegraphics[width=\textwidth]{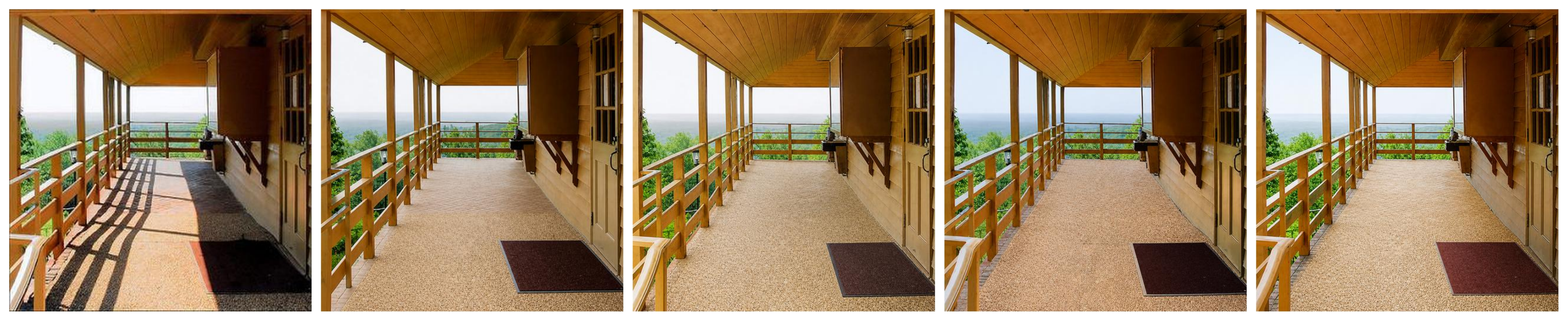}
\includegraphics[width=\textwidth]{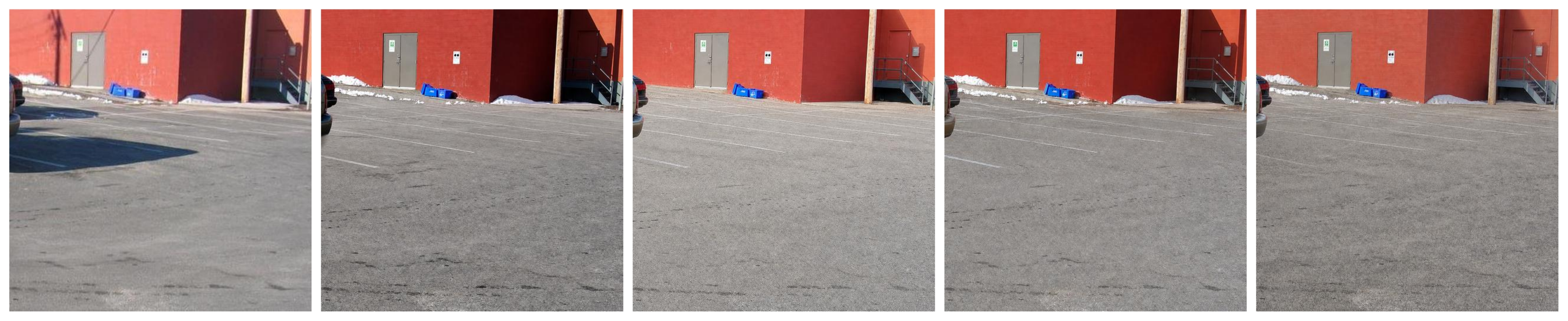}
\makebox[\textwidth]{%
    \makebox[0.20\textwidth]{\small Input}%
    \makebox[0.20\textwidth]{\small Generic}%
    \makebox[0.20\textwidth]{\small Physics cand. 1}%
    \makebox[0.20\textwidth]{\small Physics cand. 2}%
    \makebox[0.20\textwidth]{\small Visually selected}%
}
\caption{\textbf{Effect of shadow-formation grounding and visual candidate selection.}
Physics-motivated prompting reduces task-level errors and produces stronger candidate edits than generic prompting. All three physics-motivated candidates are shown, with the visually selected output in the final column.}
\label{fig:supp_grounding_selection}
\end{figure*}

\paragraph{Shadow-formation grounding and visual selection.}
\Figref{fig:supp_grounding_selection} compares generic outputs with physics-motivated candidates. Generic prompting can leave residual shadows, introduce boundary artifacts, or reinterpret dark regions as material or object structure. Shadow-formation grounding produces a stronger candidate pool, and the visual selector chooses the candidate that best balances complete shadow removal with scene, material, and structural consistency. The examples also illustrate why comparing multiple stochastic edits is more reliable than accepting a single output.

\begin{figure*}[!t]
\centering
\includegraphics[width=\linewidth]{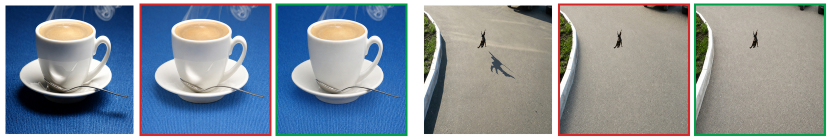}
\makebox[\linewidth][c]{%
    \makebox[0.163\linewidth][c]{\small Input}%
    \makebox[0.163\linewidth][c]{\small Visual}%
    \makebox[0.163\linewidth][c]{\small Final}%
    \hspace{0.022\linewidth}%
    \makebox[0.163\linewidth][c]{\small Input}%
    \makebox[0.163\linewidth][c]{\small Visual}%
    \makebox[0.163\linewidth][c]{\small Final}%
}
\caption{\textbf{Effect of Retinex-LoG-supported selection.}
Red frames indicate candidates selected through visual assessment alone, while green frames indicate the final selections obtained with Retinex-LoG structural evidence. The additional evidence helps reject candidates with residual shadows and select candidates that better preserve scene structure.}
\label{fig:supp_retinex_selection}
\end{figure*}

\paragraph{Retinex-LoG-supported selection.}
\Figref{fig:supp_retinex_selection} compares visual-only selection with the final selection supported by Retinex-LoG evidence. Although both candidates may appear plausible, visual assessment alone can favor incomplete shadow removal. By exposing residual, lost, and added local structure after suppressing low-frequency illumination, Retinex-LoG helps the selector achieve more complete removal while preserving the underlying texture.

\begin{figure*}[!t]
\centering
\includegraphics[width=\linewidth]{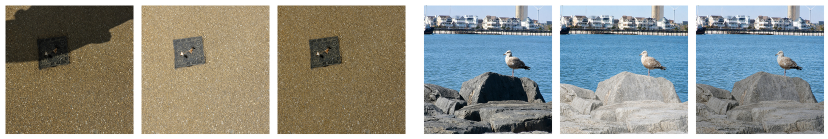}
\makebox[\linewidth][c]{%
    \makebox[0.163\linewidth][c]{\small Input}%
    \makebox[0.163\linewidth][c]{\small Selected}%
    \makebox[0.163\linewidth][c]{\small Corrected}%
    \hspace{0.022\linewidth}%
    \makebox[0.163\linewidth][c]{\small Input}%
    \makebox[0.163\linewidth][c]{\small Selected}%
    \makebox[0.163\linewidth][c]{\small Corrected}%
}
\caption{\textbf{Effect of segmentation-guided regional correction.}
Selected proposals remove the dominant shadows but can retain spatially varying brightness or color shifts. Regional correction improves agreement with the surrounding scene while preserving image structure.}
\label{fig:supp_regional_correction}
\end{figure*}

\paragraph{Segmentation-guided regional correction.}
\Figref{fig:supp_regional_correction} shows the input, selected proposal, and regionally corrected output. Candidate selection removes the dominant shadow, but the proposal may retain broad or region-specific brightness and chrominance shifts. We partition the proposal into class-agnostic regions using CropFormer~\citep{qi2023high}. For each region, the correction stage estimates offsets from reliable pixels outside the dilated shadow mask, improving local color consistency without reintroducing shadow structure from the input.

\subsection{Failure Cases and Limitations}
\label{sec:supp_construction_limitations}

\begin{figure*}[!t]
\centering
\includegraphics[width=\linewidth]{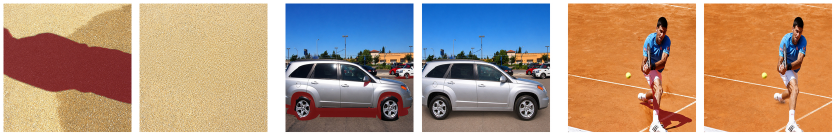}
\makebox[\linewidth][c]{%
    \makebox[0.16\linewidth][c]{\small Input + mask}%
    \makebox[0.16\linewidth][c]{\small Target}%
    \hspace{0.02\linewidth}%
    \makebox[0.16\linewidth][c]{\small Input + mask}%
    \makebox[0.16\linewidth][c]{\small Target}%
    \hspace{0.02\linewidth}%
    \makebox[0.16\linewidth][c]{\small Input + mask}%
    \makebox[0.16\linewidth][c]{\small Target}%
}
\caption{\textbf{Representative construction failures.}
Each pair shows the mask-overlay input and constructed target. The first example exhibits texture smoothing and loss of valid scene detail. The second illustrates the recurring residual-shadow problem beneath vehicles. The third shows incomplete removal of an attached shadow.}
\label{fig:supp_failure_cases}
\end{figure*}

\paragraph{Failure cases.}
Although the evaluation, selection, and correction stages reduce major construction errors, several failure modes remain. As shown in \Figref{fig:supp_failure_cases}, generative editing can smooth fine texture or remove valid scene detail. Residual shadows beneath vehicles represent a recurring failure mode. Attached shadows may also be only partially removed.

\paragraph{Limitations.}
CDD measures color consistency across the shadow boundary but does not capture changes elsewhere in the image. We therefore additionally measure RGB RMSE between the input and output over non-shadow pixels on the normalized SRR benchmark, using the 0--255 scale. Averaged over two independent runs, the final workflow obtains a mean RMSE of $23.13$, compared with $3.44$--$18.39$ for the evaluated shadow removal methods~\citep{le2019shadowdecomposition,guo2023shadowformer,guo2023shadowdiffusion,li2023inpainting,hu2025shadowrefine}. Unlike methods that largely preserve non-shadow pixels, a general-purpose editor regenerates the image and may not remain perfectly aligned with the input at the pixel level. The relatively high RMSE of ShadowDiffusion ($18.39$) is consistent with this effect. Regional correction reduces the mean non-shadow RMSE of the Retinex-LoG-selected outputs from $25.48$ to $23.13$, but the remaining gap reflects both pixel-level misalignment and unintended changes outside the target region, including color shifts, texture smoothing, and subtle structural alterations.
\section{Implementation and Evaluation Details}
\label{sec:supp_implementation}

\subsection{Baseline Checkpoints and Evaluation Protocol}
\label{sec:supp_evaluation_protocol}

\Tabref{tab:supp_baseline_protocol} records the exact checkpoint and inference settings used for evaluation. We evaluate all 540 ISTD+ and 3,021 AgenticShadow test images, restoring every prediction to its target resolution before computing metrics.
PSNR, SSIM, and MAE are computed in RGB space with values in $[0,255]$; SSIM is averaged across the three color channels. LAB RMSE is computed after converting the prediction and target from sRGB to LAB. We first average each metric over images within each domain, then compute the AgenticShadow macro average as the unweighted mean of the four domain-level results. ISTD+ is reported separately.

\begin{table*}[!t]
\centering
\caption{\textbf{Checkpoint and inference settings.}
Combined denotes training with ISTD+ and AgenticShadow, while ISTD+ / INS~\citep{xu2024omnisr} denotes separate official checkpoints. Outputs are restored to the target resolution before evaluation.}
\label{tab:supp_baseline_protocol}
\small
\setlength{\tabcolsep}{3.5pt}
\renewcommand{\arraystretch}{1.14}
\begin{tabular*}{\textwidth}{@{\extracolsep{\fill}}llcl@{}}
\toprule
Method & Checkpoint condition & Mask & Inference \\
\midrule

SID~\citep{le2019shadowdecomposition}
& Official ISTD+ & Yes & Resize to $512\times512$ \\
\midrule

ShadowFormer~\citep{guo2023shadowformer}
& Official ISTD+ & Yes & Resize to $640\times480$ \\
\midrule

Inpaint4Shadow~\citep{li2023inpainting}
& Official ISTD+ & Yes & Resize to $256\times256$ \\
\midrule

\multirow[c]{2}{*}{StableSR~\citep{xu2025detail}}
& \multirow[c]{2}{*}{Official ISTD+ / INS}
& \multirow[c]{2}{*}{No}
& Native resolution; 20 steps \\
& & & Padded to a multiple of 16 \\
\midrule

\multirow[c]{2}{*}{ShadowDiffusion~\citep{guo2023shadowdiffusion}}
& Official ISTD+
& \multirow[c]{2}{*}{Yes}
& \multirow[c]{2}{*}{\shortstack[l]{Full $512\times512$ inference\\25 DDIM steps}} \\
& Combined E400 & & \\
\midrule

\multirow[c]{2}{*}{HomoFormer~\citep{xiao2024homoformer}}
& Official ISTD+
& \multirow[c]{2}{*}{Yes}
& \multirow[c]{2}{*}{\shortstack[l]{$384\times384$ tiles\\30-pixel overlap}} \\
& Combined E200 & & \\
\midrule

\multirow[c]{2}{*}{PhaSR~\citep{lee2026phasr}}
& Official ISTD+ / INS
& \multirow[c]{2}{*}{No}
& \multirow[c]{2}{*}{\shortstack[l]{Native resolution\\Padded to a multiple of 128}} \\
& Combined E200 & & \\
\bottomrule
\end{tabular*}
\end{table*}

\subsection{PF and PF-MF Implementation and Training}
\label{sec:supp_pf_implementation}

Combined-data training uses the natural concatenation of 1,330 ISTD+ and 14,117 AgenticShadow examples without source reweighting. Image, mask, and target triplets receive the same random $384\times384$ crop and synchronized horizontal-flip and $90^\circ$-rotation augmentations. The Charbonnier loss~\citep{charbonnier1994two} uses $\epsilon=10^{-3}$.

The component analysis compares semantic-only, depth-only, and full PF variants after 50 additional epochs from the same HomoFormer E100 checkpoint. The final PF model is initialized from HomoFormer E100, while PF-MF transfers the compatible RGB weights from the same checkpoint. Both models are then trained for 100 epochs; their backbones therefore receive 200 cumulative epochs of combined-data training, while the newly added prior modules are trained for 100 epochs. For the dataset-prior analysis, the PF restoration network is trained from scratch on ISTD+ for 400 epochs following the HomoFormer training regimen, while both prior extractors remain frozen.

\subsection{Model Complexity and Runtime}
\label{sec:supp_efficiency}

We measure efficiency on one NVIDIA RTX A6000 at batch size 1 over 40 fixed $512\times512$ test images, ten from each AgenticShadow domain. Each model processes five warm-up images before timing. Mean time per image includes model-specific preprocessing, online prior extraction, tiled inference and stitching where applicable, and output conversion, but excludes model loading and disk I/O. Peak VRAM is the maximum allocated GPU memory during measurement.

As shown in \Tabref{tab:supp_efficiency}, PF and PF-MF require approximately one second per image. Their larger online parameter counts and memory usage arise primarily from running both frozen prior extractors, while their trainable parameter counts remain close to those of HomoFormer and PhaSR.

\begin{table*}[!t]
\centering
\caption{\textbf{Model complexity and inference efficiency.}
We compare ShadowDiffusion~\citep{guo2023shadowdiffusion}, HomoFormer~\citep{xiao2024homoformer}, PhaSR~\citep{lee2026phasr}, PF, and PF-MF. Online parameters include all networks executed during inference. PhaSR uses precomputed geometry, whereas PF and PF-MF execute both frozen prior extractors online. Runtime and peak VRAM are measured on an NVIDIA RTX A6000 during end-to-end processing of $512\times512$ images.}
\label{tab:supp_efficiency}
\small
\setlength{\tabcolsep}{5pt}
\renewcommand{\arraystretch}{1.1}
\begin{tabular*}{\textwidth}{@{\extracolsep{\fill}}lcccc@{}}
\toprule
Method & Online params (M) & Trainable params (M) & Time/image (s) & Peak VRAM (GiB) \\
\midrule
ShadowDiffusion & 55.52  & 55.52 & 2.558 & 1.51 \\
HomoFormer      & 17.81  & 17.81 & 0.505 & 0.92 \\
PhaSR           & 323.37 & 19.00 & 0.445 & 2.80 \\
PF              & 658.93 & 19.25 & 0.999 & 3.33 \\
PF-MF           & 658.61 & 18.92 & 0.978 & 3.32 \\
\bottomrule
\end{tabular*}
\end{table*}

\section{Additional Results}
\label{sec:supp_additional_results}

\subsection{Full Per-Domain Evaluation}
\label{sec:supp_per_domain}

\begin{table*}[!t]
\centering
\caption{\textbf{Full evaluation on the SBU and CUHK-Shadow domains.}
Official ISTD+ checkpoints are evaluated without adaptation, while $^\dagger$ denotes training with ISTD+ and AgenticShadow. Best and second-best results are \textbf{bolded} and \underline{underlined}, respectively.}
\label{tab:supp_sbu_cuhk}
\small
\setlength{\tabcolsep}{3.2pt}
\renewcommand{\arraystretch}{1.06}
\begin{tabular*}{\textwidth}{@{\extracolsep{\fill}}lcccccccc@{}}
\toprule
& \multicolumn{4}{c}{SBU}
& \multicolumn{4}{c}{CUHK-Shadow} \\
\cmidrule(lr){2-5}\cmidrule(lr){6-9}
Method
& PSNR$\uparrow$ & SSIM$\uparrow$ & MAE$\downarrow$ & LAB$\downarrow$
& PSNR$\uparrow$ & SSIM$\uparrow$ & MAE$\downarrow$ & LAB$\downarrow$ \\
\midrule
\multicolumn{9}{l}{\textit{Official ISTD+ checkpoints}} \\
SID~\citep{le2019shadowdecomposition}
& 20.28 & 0.6258 & 18.26 & 7.25
& 16.32 & 0.5643 & 30.43 & 10.86 \\
ShadowFormer~\citep{guo2023shadowformer}
& 21.02 & 0.6506 & 16.99 & 6.63
& 17.09 & 0.6008 & 28.50 & 10.01 \\
Inpaint4Shadow~\citep{li2023inpainting}
& 20.51 & 0.5883 & 18.30 & 7.12
& 17.54 & 0.5406 & 26.43 & 9.72 \\
StableSR~\citep{xu2025detail}
& 20.34 & 0.6345 & 18.83 & 7.32
& 18.38 & 0.6173 & 23.94 & 8.84 \\
ShadowDiffusion~\citep{guo2023shadowdiffusion}
& 20.75 & 0.6401 & 17.73 & 6.83
& 17.88 & 0.6026 & 25.22 & 9.09 \\
HomoFormer~\citep{xiao2024homoformer}
& 20.85 & 0.6430 & 17.41 & 6.78
& 18.01 & 0.6085 & 24.69 & 9.08 \\
PhaSR~\citep{lee2026phasr}
& 20.28 & 0.6405 & 18.72 & 7.32
& 18.47 & 0.6240 & 23.44 & 8.71 \\
\midrule
\multicolumn{9}{l}{\textit{With AgenticShadow supervision}} \\
ShadowDiffusion$^\dagger$
& 21.85 & 0.6578 & 16.12 & 5.84
& 19.80 & 0.6618 & 19.96 & 7.40 \\
HomoFormer$^\dagger$
& 23.61 & 0.6783 & 12.79 & 4.90
& 21.53 & 0.6926 & 15.89 & 6.23 \\
PhaSR$^\dagger$
& 23.66 & \textbf{0.6924} & 12.92 & 4.92
& 21.62 & \textbf{0.7073} & 15.93 & 6.19 \\
PF
& \textbf{23.93} & 0.6888 & \underline{12.39} & \textbf{4.73}
& \textbf{21.80} & 0.7030 & \textbf{15.38} & \textbf{6.05} \\
PF-MF
& \underline{23.90} & \underline{0.6896} & \textbf{12.38} & \underline{4.75}
& \underline{21.72} & \underline{0.7039} & \underline{15.61} & \underline{6.11} \\
\bottomrule
\end{tabular*}
\end{table*}

\begin{table*}[!t]
\centering
\caption{\textbf{Full evaluation on the ASFW and S-EO domains.}
Official ISTD+ checkpoints are evaluated without adaptation, while $^\dagger$ denotes training with ISTD+ and AgenticShadow. Best and second-best results are \textbf{bolded} and \underline{underlined}, respectively.}
\label{tab:supp_asfw_seo}
\small
\setlength{\tabcolsep}{3.2pt}
\renewcommand{\arraystretch}{1.06}
\begin{tabular*}{\textwidth}{@{\extracolsep{\fill}}lcccccccc@{}}
\toprule
& \multicolumn{4}{c}{ASFW}
& \multicolumn{4}{c}{S-EO} \\
\cmidrule(lr){2-5}\cmidrule(lr){6-9}
Method
& PSNR$\uparrow$ & SSIM$\uparrow$ & MAE$\downarrow$ & LAB$\downarrow$
& PSNR$\uparrow$ & SSIM$\uparrow$ & MAE$\downarrow$ & LAB$\downarrow$ \\
\midrule
\multicolumn{9}{l}{\textit{Official ISTD+ checkpoints}} \\
SID~\citep{le2019shadowdecomposition}
& 20.02 & 0.7687 & 19.38 & 7.48
& 16.50 & 0.6027 & 27.02 & 10.23 \\
ShadowFormer~\citep{guo2023shadowformer}
& 20.18 & 0.7801 & 19.36 & 7.35
& 16.01 & 0.6052 & 28.57 & 10.74 \\
Inpaint4Shadow~\citep{li2023inpainting}
& 19.90 & 0.7523 & 19.88 & 7.58
& 16.43 & 0.5665 & 27.62 & 10.08 \\
StableSR~\citep{xu2025detail}
& 17.14 & 0.7367 & 28.16 & 10.01
& 15.75 & 0.6140 & 29.98 & 10.93 \\
ShadowDiffusion~\citep{guo2023shadowdiffusion}
& 20.10 & 0.7704 & 19.61 & 7.47
& 16.17 & 0.6119 & 28.12 & 10.50 \\
HomoFormer~\citep{xiao2024homoformer}
& 20.57 & 0.7789 & 18.54 & 7.00
& 15.95 & 0.6083 & 28.77 & 10.87 \\
PhaSR~\citep{lee2026phasr}
& 17.00 & 0.7454 & 28.03 & 10.36
& 15.87 & 0.6125 & 28.82 & 10.92 \\
\midrule
\multicolumn{9}{l}{\textit{With AgenticShadow supervision}} \\
ShadowDiffusion$^\dagger$
& 21.81 & 0.8030 & 15.79 & 5.97
& 18.40 & 0.6583 & 22.46 & 7.98 \\
HomoFormer$^\dagger$
& 23.03 & 0.8177 & 13.37 & 5.33
& 19.65 & 0.6839 & 18.95 & 7.02 \\
PhaSR$^\dagger$
& 23.11 & 0.8240 & 14.29 & 5.31
& 19.87 & \textbf{0.7000} & 18.63 & 6.90 \\
PF
& \textbf{24.33} & \textbf{0.8348} & \textbf{11.59} & \textbf{4.67}
& \underline{19.96} & \underline{0.6990} & \textbf{18.08} & \textbf{6.76} \\
PF-MF
& \underline{23.81} & \underline{0.8298} & \underline{12.55} & \underline{4.97}
& \textbf{19.99} & 0.6979 & \underline{18.14} & \underline{6.78} \\
\bottomrule
\end{tabular*}
\end{table*}

\Tabref{tab:supp_sbu_cuhk} and \Tabref{tab:supp_asfw_seo} provide the complete per-domain results underlying the macro averages in the main paper. AgenticShadow supervision improves all four metrics for each retrained architecture in every domain, demonstrating consistent benefits across general, facial, and remote-sensing scenes. PF and PF-MF rank first or second in PSNR, MAE, and LAB RMSE across all four domains, with PF-MF remaining close despite requiring no input mask, while PhaSR$^\dagger$ achieves the highest SSIM on SBU, CUHK-Shadow, and S-EO.

\subsection{Cross-Dataset and Prior Analysis}
\label{sec:supp_cross_dataset}

\begin{table*}[!t]
\centering
\caption{\textbf{Synthetic-to-real transfer.}
We compare official ISTD+ and INS~\citep{xu2024omnisr} checkpoints of
StableSR~\citep{xu2025detail} and PhaSR~\citep{lee2026phasr}.
AgenticShadow results are four-domain macro averages.}
\label{tab:supp_synthetic_transfer}
\small
\setlength{\tabcolsep}{3.5pt}
\renewcommand{\arraystretch}{1.08}
\begin{tabular*}{\textwidth}{@{\extracolsep{\fill}}llcccccccc@{}}
\toprule
& & \multicolumn{4}{c}{ISTD+}
& \multicolumn{4}{c}{AgenticShadow} \\
\cmidrule(lr){3-6}\cmidrule(lr){7-10}
Model & Checkpoint
& PSNR$\uparrow$ & SSIM$\uparrow$ & MAE$\downarrow$ & LAB$\downarrow$
& PSNR$\uparrow$ & SSIM$\uparrow$ & MAE$\downarrow$ & LAB$\downarrow$ \\
\midrule
\multirow{2}{*}{StableSR}
& ISTD+ & 32.23 & 0.9272 & 4.84 & 2.03
& 17.90 & 0.6506 & 25.23 & 9.28 \\
& INS & 22.76 & 0.8986 & 12.14 & 5.82
& 16.55 & 0.6350 & 28.93 & 11.12 \\
\midrule
\multirow{2}{*}{PhaSR}
& ISTD+ & 32.12 & 0.9330 & 4.82 & 2.13
& 17.90 & 0.6556 & 24.75 & 9.33 \\
& INS & 26.03 & 0.9129 & 8.93 & 4.32
& 17.24 & 0.6415 & 26.90 & 10.37 \\
\bottomrule
\end{tabular*}
\end{table*}

\paragraph{Synthetic-to-real transfer.}
Large-scale synthetic data provide another route to paired supervision: OmniSR~\citep{xu2024omnisr} introduced INS with over 30,000 rendered shadow and shadow-free image pairs spanning diverse objects and direct and indirect illumination. We therefore evaluate official INS checkpoints of StableSR and PhaSR to test whether this scale transfers to real imagery. As shown in \Tabref{tab:supp_synthetic_transfer}, both INS checkpoints underperform their ISTD+ counterparts across all metrics on ISTD+ and AgenticShadow, revealing a persistent synthetic-to-real gap and underscoring the importance of large-scale supervision constructed from diverse real-world images.

\begin{table*}[!t]
\centering
\caption{\textbf{Effect of HomoFormer training data.}
The AgenticShadow and Combined conditions are initialized from the official
ISTD+ checkpoint; Combined denotes training with ISTD+ and AgenticShadow.
AgenticShadow results are four-domain macro averages.}
\label{tab:supp_homo_training}
\small
\setlength{\tabcolsep}{7pt}
\renewcommand{\arraystretch}{1.08}
\begin{tabular*}{\textwidth}{@{\extracolsep{\fill}}lcccccc@{}}
\toprule
& \multicolumn{2}{c}{ISTD+}
& \multicolumn{4}{c}{AgenticShadow} \\
\cmidrule(lr){2-3}\cmidrule(lr){4-7}
Training data
& PSNR$\uparrow$ & LAB$\downarrow$
& PSNR$\uparrow$ & SSIM$\uparrow$ & MAE$\downarrow$ & LAB$\downarrow$ \\
\midrule
ISTD+
& \underline{32.53} & \underline{1.92}
& 18.85 & 0.6597 & 22.36 & 8.43 \\
AgenticShadow
& 27.96 & 2.74
& \underline{21.85} & \underline{0.7152}
& \underline{15.45} & \underline{5.94} \\
Combined
& \textbf{32.92} & \textbf{1.85}
& \textbf{21.95} & \textbf{0.7181}
& \textbf{15.25} & \textbf{5.87} \\
\bottomrule
\end{tabular*}
\end{table*}

\paragraph{Complementary paired supervision.}
\Tabref{tab:supp_homo_training} examines how controlled and constructed pairs complement one another. Training with only AgenticShadow after ISTD+ initialization reduces macro LAB RMSE from 8.43 to 5.94 but increases ISTD+ LAB RMSE from 1.92 to 2.74, indicating that controlled captures remain a useful anchor for photometric fidelity. Combined training performs best on both evaluations, showing that AgenticShadow adds substantial real-world diversity while controlled pairs help preserve precise restoration behavior.

\paragraph{Prior components.}
\Tabref{tab:supp_pf_components} shows that semantic and depth priors independently improve every AgenticShadow domain, with semantic information providing the larger gain. Combining both priors performs best on all four domains and the macro average, while ISTD+ performance remains nearly unchanged across the matched variants. This indicates that prior fusion is most beneficial under the broader scene and shadow diversity of AgenticShadow.

\begin{table*}[!t]
\centering
\caption{\textbf{Per-domain PF component analysis.}
All values are LAB RMSE. Each variant is initialized from HomoFormer E100 and
trained for 50 additional epochs; Macro is the unweighted average of the four
AgenticShadow domains.}
\label{tab:supp_pf_components}
\small
\setlength{\tabcolsep}{8pt}
\renewcommand{\arraystretch}{1.08}
\begin{tabular*}{\textwidth}{@{\extracolsep{\fill}}lcccccc@{}}
\toprule
Variant & ISTD+ & SBU & CUHK & ASFW & S-EO & Macro \\
\midrule
HomoFormer
& 1.860 & 4.907 & 6.266 & 5.434 & 7.041 & 5.912 \\
Depth only
& 1.859 & 4.835 & 6.155 & 5.122 & 6.964 & 5.769 \\
Semantic only
& \textbf{1.857} & 4.731 & 6.051 & 4.750 & 6.807 & 5.585 \\
PF
& 1.861 & \textbf{4.723} & \textbf{6.045}
& \textbf{4.737} & \textbf{6.804} & \textbf{5.577} \\
\bottomrule
\end{tabular*}
\end{table*}

\subsection{Qualitative Results and Videos}
\label{sec:supp_qualitative}

\paragraph{Videos.}
We additionally evaluate video sequences, comparing PF with pretrained HomoFormer~\citep{xiao2024homoformer} on ViSha~\citep{chen2021triple} using the provided masks, and PF-MF with SP+M+I-Net~\citep{le2022physics} on SBU-TimeLapse without input masks. All methods are applied independently to individual frames without temporal modeling. Although our baselines are not optimized for sequential data, PF and PF-MF produce cleaner and more complete shadow removal than the compared methods across the videos.

\begin{figure*}[!t]
\centering
\includegraphics[width=\textwidth]{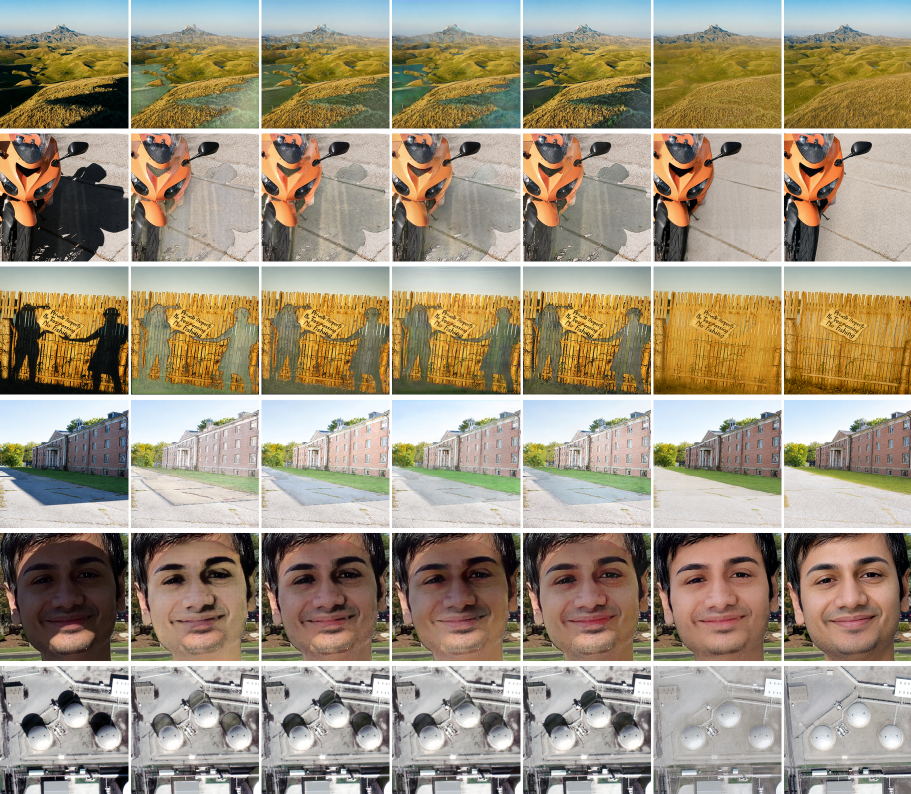}

\makebox[\textwidth][c]{%
    \makebox[0.1405\textwidth][c]{\scriptsize Input}%
    \hspace{0.0027\textwidth}%
    \makebox[0.1405\textwidth][c]{\scriptsize SID}%
    \hspace{0.0027\textwidth}%
    \makebox[0.1405\textwidth][c]{\scriptsize ShadowFormer}%
    \hspace{0.0027\textwidth}%
    \makebox[0.1405\textwidth][c]{\scriptsize Inpaint4Shadow}%
    \hspace{0.0027\textwidth}%
    \makebox[0.1405\textwidth][c]{\scriptsize ShadowDiffusion}%
    \hspace{0.0027\textwidth}%
    \makebox[0.1405\textwidth][c]{\scriptsize PF}%
    \hspace{0.0027\textwidth}%
    \makebox[0.1405\textwidth][c]{\scriptsize Target}%
}

\caption{\textbf{Qualitative comparison with official ISTD+ checkpoints.}
We compare SID~\citep{le2019shadowdecomposition}, ShadowFormer~\citep{guo2023shadowformer}, Inpaint4Shadow~\citep{li2023inpainting}, and ShadowDiffusion~\citep{guo2023shadowdiffusion} with PF across the four AgenticShadow domains. The official checkpoints frequently leave residual shadows or introduce color shifts, whereas PF produces more complete removal and outputs closer to the constructed targets.}
\label{fig:supp_qualitative_official}
\end{figure*}

\begin{figure*}[!t]
\centering
\includegraphics[width=\textwidth]{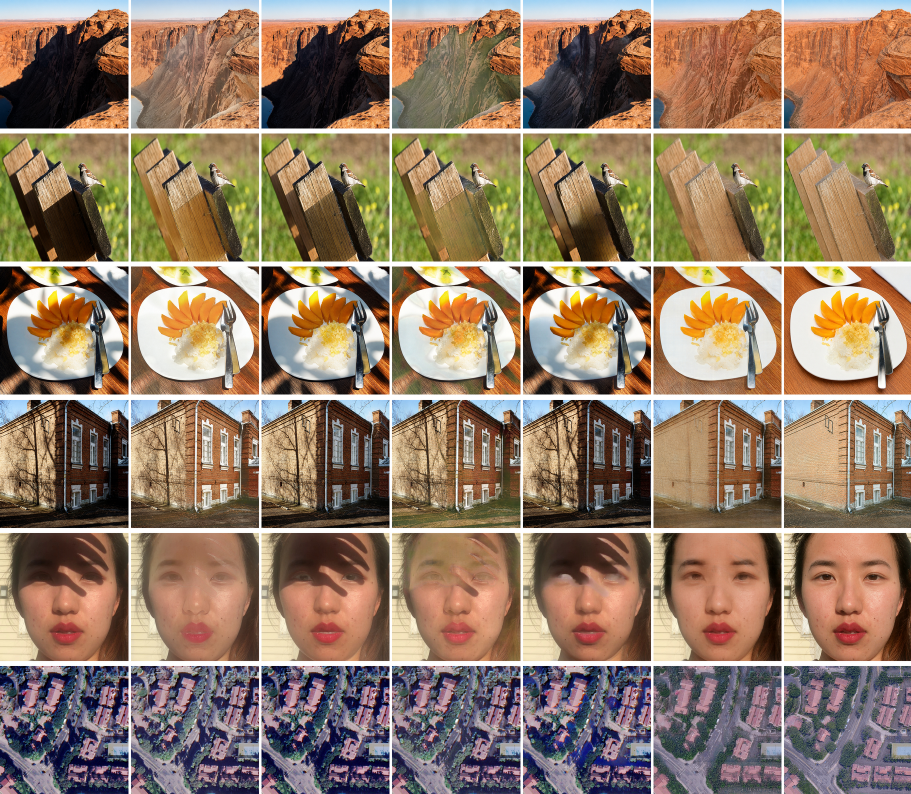}

\makebox[\textwidth][c]{%
    \makebox[0.1405\textwidth][c]{\scriptsize Input}%
    \hspace{0.0027\textwidth}%
    \makebox[0.1405\textwidth][c]{\scriptsize StableSR-ISTD+}%
    \hspace{0.0027\textwidth}%
    \makebox[0.1405\textwidth][c]{\scriptsize StableSR-INS}%
    \hspace{0.0027\textwidth}%
    \makebox[0.1405\textwidth][c]{\scriptsize PhaSR-ISTD+}%
    \hspace{0.0027\textwidth}%
    \makebox[0.1405\textwidth][c]{\scriptsize PhaSR-INS}%
    \hspace{0.0027\textwidth}%
    \makebox[0.1405\textwidth][c]{\scriptsize PF}%
    \hspace{0.0027\textwidth}%
    \makebox[0.1405\textwidth][c]{\scriptsize Target}%
}

\caption{\textbf{Qualitative comparison of ISTD+ and INS checkpoints.}
We compare the official ISTD+ and synthetic INS~\citep{xu2024omnisr} checkpoints of StableSR~\citep{xu2025detail} and PhaSR~\citep{lee2026phasr} with PF. The INS-trained models continue to exhibit residual shadows and appearance shifts on real-world AgenticShadow scenes, illustrating that large-scale synthetic supervision does not eliminate the synthetic-to-real gap.}
\label{fig:supp_qualitative_synthetic}
\end{figure*}

\begin{figure*}[!t]
\centering
\includegraphics[width=\textwidth]{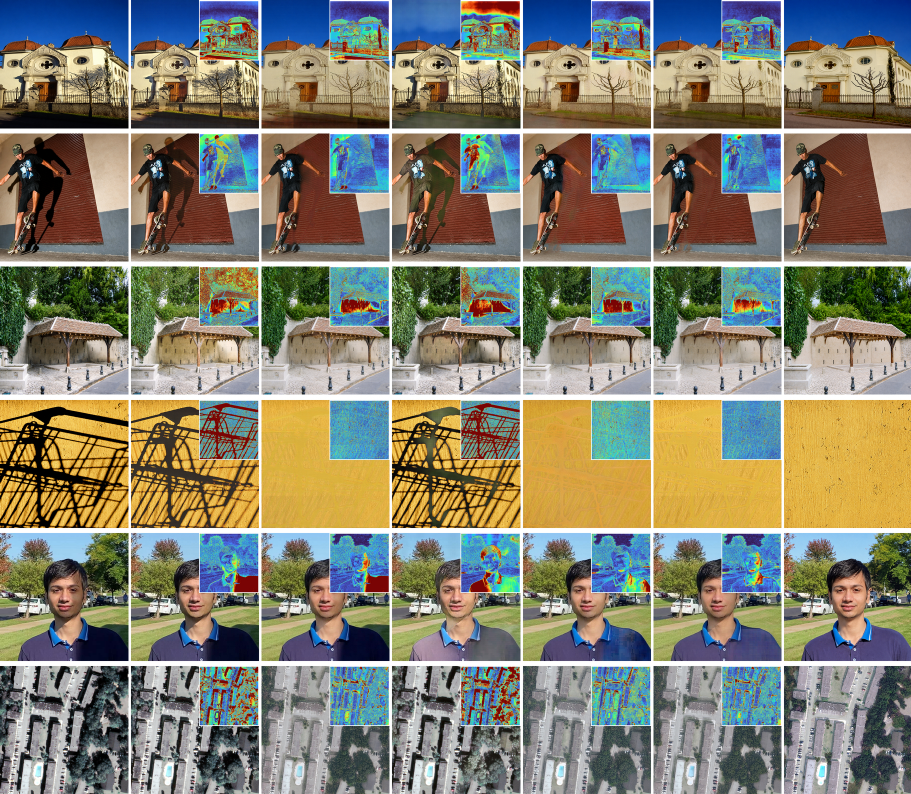}

\makebox[\textwidth][c]{%
    \makebox[0.1405\textwidth][c]{\scriptsize Input}%
    \hspace{0.0027\textwidth}%
    \makebox[0.1405\textwidth][c]{\scriptsize HomoFormer}%
    \hspace{0.0027\textwidth}%
    \makebox[0.1405\textwidth][c]{\scriptsize HomoFormer$^\dagger$}%
    \hspace{0.0027\textwidth}%
    \makebox[0.1405\textwidth][c]{\scriptsize PhaSR}%
    \hspace{0.0027\textwidth}%
    \makebox[0.1405\textwidth][c]{\scriptsize PhaSR$^\dagger$}%
    \hspace{0.0027\textwidth}%
    \makebox[0.1405\textwidth][c]{\scriptsize PF-MF}%
    \hspace{0.0027\textwidth}%
    \makebox[0.1405\textwidth][c]{\scriptsize Target}%
}

\caption{\textbf{Additional qualitative comparison of AgenticShadow supervision.}
We compare the official ISTD+ checkpoints of HomoFormer~\citep{xiao2024homoformer} and PhaSR~\citep{lee2026phasr}, their counterparts trained with ISTD+ and AgenticShadow ($^\dagger$), and PF-MF. AgenticShadow supervision produces cleaner and more complete shadow removal, while PF-MF achieves strong results without input masks. Colored insets show mean absolute RGB error using a shared scale.}
\label{fig:supp_qualitative_expanded}
\end{figure*}

\paragraph{Additional qualitative comparisons.}
We provide three complementary comparisons:
(1) \Figref{fig:supp_qualitative_official} completes qualitative coverage of the official ISTD+ checkpoints;
(2) \Figref{fig:supp_qualitative_synthetic} compares ISTD+ and INS checkpoints to illustrate the synthetic-to-real generalization gap; and
(3) \Figref{fig:supp_qualitative_expanded} contrasts ISTD+-pretrained models with their AgenticShadow-trained counterparts and PF-MF, demonstrating the benefit of broader paired supervision and the effectiveness of the mask-free variant.

\end{document}